\documentclass{article}

\usepackage{arxiv}

\usepackage[utf8]{inputenc} % allow utf-8 input
\usepackage[T1]{fontenc}    % use 8-bit T1 fonts
\usepackage{hyperref}       % hyperlinks
\usepackage{url}            % simple URL typesetting
\usepackage{booktabs}       % professional-quality tables
\usepackage{amsfonts}       % blackboard math symbols
\usepackage{nicefrac}       % compact symbols for 1/2, etc.
\usepackage{microtype}      % microtypography
\usepackage{lipsum}
\usepackage{graphicx}
\title{Semantic scene synthesis: Application to assistive systems}

\author{
  Chayma Zatout \\
  RIIMA Laboratory\\Computer Science Department\\
  USTHB University\\
  \texttt{czatout@usthb.dz} \\
   \And
 Slimane Larabi \\
  RIIMA Laboratory\\Computer Science Department\\
  USTHB University\\
  \texttt{slarabi@usthb.dz} \\
}

\begin{document}
\maketitle

\begin{abstract}
The aim of this work is to provide a semantic scene synthesis from a single depth image. This is used in assistive aid systems for visually impaired and blind people that allows them to understand their surroundings by the touch sense. The fact that blind people use touch to recognize objects and rely
on listening to replace sight, motivated us to propose this work. \\
First, the acquired depth image is segmented and each segment is classified in the context of assistive systems using a deep learning network. Second, inspired by the Braille system and the Japanese writing system Kanji, the obtained classes are coded with semantic labels. The scene is then synthesized using these labels and the extracted geometric features.\\
Our system is able to predict more than 17 classes only by understanding the
provided illustrative labels. For the remaining objects, their geometric features
are transmitted. The labels and the geometric features are mapped on a synthesis area to be sensed by the touch sense.\\
Experiments are conducted on noisy and incomplete data including acquired depth images of indoor scenes and public datasets. The obtained results are reported and discussed.
\end{abstract}

% keywords can be removed

\section{Introduction}\label{sec:introduction}
In order to accomplish daily tasks, people involve their five senses, namely sight, hearing, taste, smell and touch. Being deprived of one of these senses will complicate the process of a given task; it will reduce human autonomy, independence and even privacy; the visually impaired find difficulties in their daily life.

With the limitations of the classical aid systems such as white canes, guide dogs and personal assistants; and with the evolution of technology, many commercial and noncommercial aid systems were proposed in the last decades. Generally, these latter rely on image processing, artificial intelligence techniques and external sensors in order to offer help for the visually impaired and blind people to improve their independence in many applications.

To transmit instructions, scene description or any other generated output, most of the assistive systems use audio-based or vibration-based output devices. It turns out that these latter hold hearing and are not too informative. Hence, the necessity of providing a semantic labeling for scene understanding that can be exploited by the touch sense. 

In this work, we propose a framework for semantic scene synthesis. From the depth image, the 3D scene is down-scaled and semantically mapped into a synthesis area using the computed labels and the extracted geometric features of the input point cloud. Two main modules are proposed: the classification module and the semantic labeling module. The first module is based on a deep learning architecture to classify depth image segments into seven semantic classes. The semantic labeling is inspired from Braille and Kanji systems. This latter is mapped into a touch-based synthesis area that can be used for many applications such as in assistive systems for visually impaired and blind people.

The remaining sections are structured as follows: in section \ref{works}, we present related works to objects classification, scenes understanding and semantic labeling for assistive systems. An overview of the proposed system is presented in section \ref{system}. In sections \ref{classification} and \ref{Semantic}, we describe our approaches for the classification and the semantic labeling module respectively. Conducted experiments are reported and discussed in section \ref{exp}. Finally, we conclude with the future works (section \ref{fcope}) and a conclusion (section \ref{conc}).
%%%%%%%%%%%%%%%%%%%%%%%%%%%%%%%%%%%%%%%%%%%%%%%%%%%%%%%%%%%%%%%%%%%%%%%%%%%%%%%%%%%%%%%%%%%%%%%%%%%%%%%
%						RELATED WORKS
%%%%%%%%%%%%%%%%%%%%%%%%%%%%%%%%%%%%%%%%%%%%%%%%%%%%%%%%%%%%%%%%%%%%%%%%%%%%%%%%%%%%%%%%%%%%%%%%%%%%%%%
\section{Related Works}\label{works}
Visually impaired aid systems consist of a set of techniques whose goal is to enhance the visually impaired life in different activities. These systems can be traditional like white canes, guide dogs or personal assistants; sophisticated by involving advanced technologies and computer science techniques \cite{jafri2014computer} or hybrid \cite{Leo2017}\cite{Tapu2018}\cite{cardin2007wearable}. The sophisticated systems process the received data from the real-world using sensors and transform it into instructions and signs that can be understood by the visually impaired people. They use depth or RGB sensors, image processing techniques, computer vision and machine learning. In this work, we focus on scene understanding and semantic labeling.
\subsection{Scene understanding and object recognition}
Object recognition in complex scenes is still a challenging problem. In the recent work \cite{Cupec2020}, they recognized objects based on alignment of convex hulls of the detected segments in a depth image. Each computed convex hull was then compared with convex hulls of target 3D object models or their parts. This alignment is performed using the Convex Template Instance descriptor. \\
In visually impaired assistive systems, detecting objects is important in most applications. Knowing their nature will provide the ability to deduct additional information such as scene understanding, auto positioning and creating free space by moving some kinds of objects like chairs. In \cite{wang2017enabling}, using a depth camera, they proposed a wearable system based on a linear classifier to classify the point cloud features into: chairs, tables, stair-up, stair-down and walls. The obtained class is coded and mapped into a braille display. In \cite{Perez2017}, they proposed a system that uses Visual Odometry, Region-growing and Euclidean cluster extraction, and depth data to determine if the horizontal planes are a valid step of a staircase. In \cite{lin2019deep}, they implemented scene understanding using deep learning techniques. The scene is captured using an RGB-D camera and the results are displayed using an earphone and a smartphone that serves as a haptic device. They adapted FuseNet \cite{hazirbas2016fusenet} and GoogleLeNet \cite{szegedy2015going} to provide semantic segmentation and orientation instructions respectively. The semantic segmentation model was designed to predict 40 different classes; however, these latter are transmitted using an audio device. The authors in \cite{wang2014segment}, segmented the input point cloud based on the cascaded decision tree using RGB-D images. The system only classifies a given segment, whether it is the ground, a wall or a table (a horizontal plane that is not the ground). As in \cite{bai2019wearable}, they proposed a lightweight CNN architecture that is able to be executed on smartphones for traveling assistive systems. They adopted PeleeNet for object detection to cover 80 different classes using the RGB images as input.
\subsection{Semantic labeling}
In \cite{Rubio2016}, they proposed a navigation system based on RGB image processing. The system transmits the generated scene and the navigation instructions on the Senseg TM device. They used the electrostatic signs to generate codes and to form textural instructions for the visually impaired and blind people. They also used colors to encode additional visual feedback for the sighted and with low vision people.

Authors in \cite{wang2017enabling}, encoded the considered object classes using the first character of the class's name: o, c, t and the space character to represent obstacles, chairs, tables and free spaces respectively. They used a braille device to transmit the occupancy grid to the user and to deliver the object's label in the
Braille system. These codes are simple to understand, but it can be ambiguous while covering a large set of objects, especially with objects having the same first character.

In our previous work \cite{zatout2019ego}, we proposed a semantic labeling for scene understanding since after understanding the scenes' components, a human-being can accomplish other tasks such as navigating. We mapped the detected planes into cylinders having a specific height and radius. These latter are mapped into a trapezoid area at a specific location. By touching this area, the user can locate objects from the free space that allows him to have an idea about their characteristics, namely their heights and their areas. The proposed semantic labeling does not reveal the object's nature, it only gives an impression about the given object.
\subsection{Discussion and Contributions}
The state of the art systems present many limitations:\\
- In indoor navigation, for example, they provide only a global description of the captured scenes (free space and obstacles without describing their nature). In some proposed systems when objects are detected, they usually use an RGB camera as an additional input sensor \cite{jeamwatthanachai2017map}. In other works, they provide obstacle classification, but by considering only a few classes such as classifying the scene components into the floor, objects that are parallel to the floor and objects that are perpendicular to floor without considering other features such as the object's height and its occupied area.\\
- The system's output is usually transmitted by an audio device. This latter suffers from some limitations as explained in \cite{zatout2019ego} and in \cite{bai2017smart} works.\\
- The proposed semantic labelings are less informative and can generate ambiguity and do not consider some geometric features that can be helpful  \cite{wang2017enabling}.\\
- The actual aid systems are vibration-based or audio-based. For the visually impaired, the hearing replaces the sight in different tasks such as detecting some events, thus, the importance of releasing it.\\
- The visually impaired use the touch for learning geometries, recognizing objects and human faces; and even for reading.

As a consequence, our objective through this work, is to propose an end-to-end assistive system that uses only the depth image as an input and it generates labels on a tangible area. This system provides a scene description designed specially for the visually impaired: it allows them to locate objects and to recognize the geometry of the captured scene. The tangible area is used as an output device to replace the traditional output device and thus the hearing is released and the labeling is more informative.
\section{System overview}\label{system}
\begin{figure}[!t]
	\centering
	\includegraphics[width=6cm]{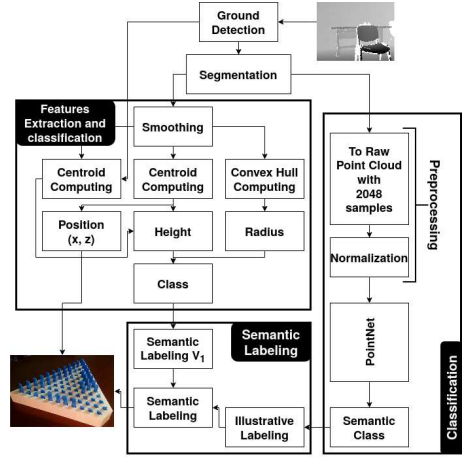}
    \caption{The three essential modules of our proposed framework: Features extraction module,
Classification module and the module that generates semantic labeling.}
	\label{fig:framwork_diag}
\end{figure}

Our proposed system (Figure \ref{fig:framwork_diag}) takes a single depth frame captured by a head-mounted depth camera as an input. After detecting the ground using the DCGD (Depth-Cut based Ground Detection) algorithm \cite{zatout2019ego}, the occupied space is extracted and segmented. Each computed segment is then fed to the deep network to perform object classification. The feature extraction module
computes geometric features for each segment such as the object's height. After that, the provided class and features are used to generate semantic labels. Finally, the captured scene is synthesized based on the features and labels computed previously. The system can be adapted to use an RGB camera as the input sensor; however, one more step that consists of estimating the depth image will be required. In the visually impaired assistive systems, knowing the distance away from objects and the geometry of the captured scene are important.
\subsection{Ground detection}
\begin{figure}[t]
	\centering
	 \includegraphics[height=2.5cm]{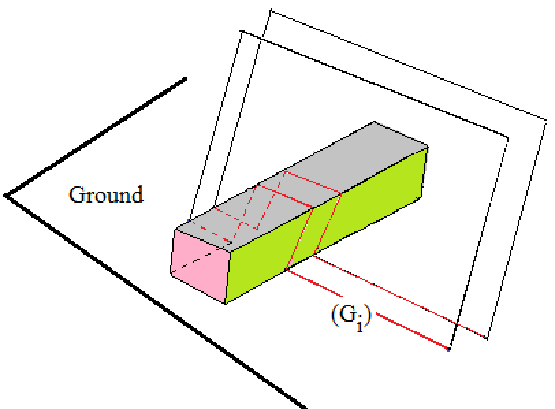}
     \includegraphics[height=2.5cm]{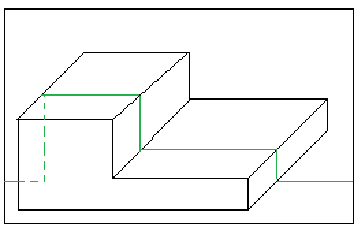}
	 \caption{(Right) The obtained curve $(G_i)$ for two cutting planes. (Left)The 3D points at a given depth with minimal $y-$coordinate (case where $xz-$plane is parallel to the ground) (colored in green). }
	 \label{fig:linedelta}
\end{figure}

%%%%%%%%%%%%%%%%%% DCGD:
The DCGD algorithm \cite{zatout2019ego} consists of computing a set of depth cuts from a given depth image. For a given depth $z_i$ , the selection on each column of the depth image pixels having $z=z_i$ will define the images of 3D points at distance $z_i$ . These 3D points define a parallel plane $(\Pi_i)$ to image plane cutting the 3D scene (Figure \ref{fig:linedelta}). The selected pixels on each column having the minimal value of the $y-component$ define the image of intersection $(G_i)$ of the plane $(\Pi_i)$ with the 3D scene. From the initial distance $z_0$ to the final distance (maximal authorized) $z_f$ , for each step $\delta z$, the curve $G_i, i=0,n$ is computed at $z=z_0+i\times \delta z$, such that $z_f=z_0+n\times \delta z$. The second step consists of removing pixels from $(G_i)$ that corresponds to objects. Each cut is divided into sub-cuts labeled either ”concave” or ”convex”. The ”convex” parts represent objects on the floor. The ground is then detected by iteratively removing the convex of the current cut to keep only concave parts which will, at the end of the process, represent the ground.
%%%%%%%%%%%%%%%%%% DCGD end.
\subsection{Point cloud segmentation}
In this work, we are interested in detecting and representing coarse information, i.e. large segments that represent salient objects. Detecting and transmitting coarse information in visually impaired assistive systems is important, knowing the scene architecture and how objects are arranged can be useful to accomplish many daily tasks such as scene understanding and navigation. Accordingly, in order to segment the obtained point cloud into coarse segments, clustering algorithms can be used for irregular object segmentation as mentioned in \cite{xie2019review}. K-means \cite{kuccak2017segmentation} \cite{aparajithan2006clustering}, mean-shift \cite{melzer2007non}, fuzzy-clustering \cite{biosca2008unsupervised} and DBSCAN \cite{czerniawski20186d}\cite{wu2018squeezeseg}\cite{wang2019improved} clustering algorithms are widely used for point cloud segmentation. 

%%%%%%%%%%%%%%%%%%%% Clustering:
K-means, mean-shift and fuzzy-clustering are centroid based algorithms: the first divides the input samples into K separate groups with equal variance while minimizing a given criterion. The mean-shift algorithm separates the input samples into blobs with smooth density. As for the fuzzy-clustering which is based on fuzzy-logic, it assigns for each sample a probability corresponding to each cluster. More the probability is high the more the sample is near to the cluster. One of the disadvantages of these algorithms is the fact that they require the number of clusters  as input which is unknown when dealing with real-world data. On the other hand, the DBSCAN algorithm seeks to separate the samples into high density clusters with low density areas. In addition, it is a non-parametric algorithm, the number of clusters does not have to be defined beforehand. Furthermore, the DBSCAN is robust to noise: during the clustering, the outliers  are detected and neglected and thus, the clustering process is not affected by them.  
%%%%%%%%%%%%%%%%%%%% Clustering end.

\begin{figure}[t]
	\centering
		\includegraphics[height=2.5cm]{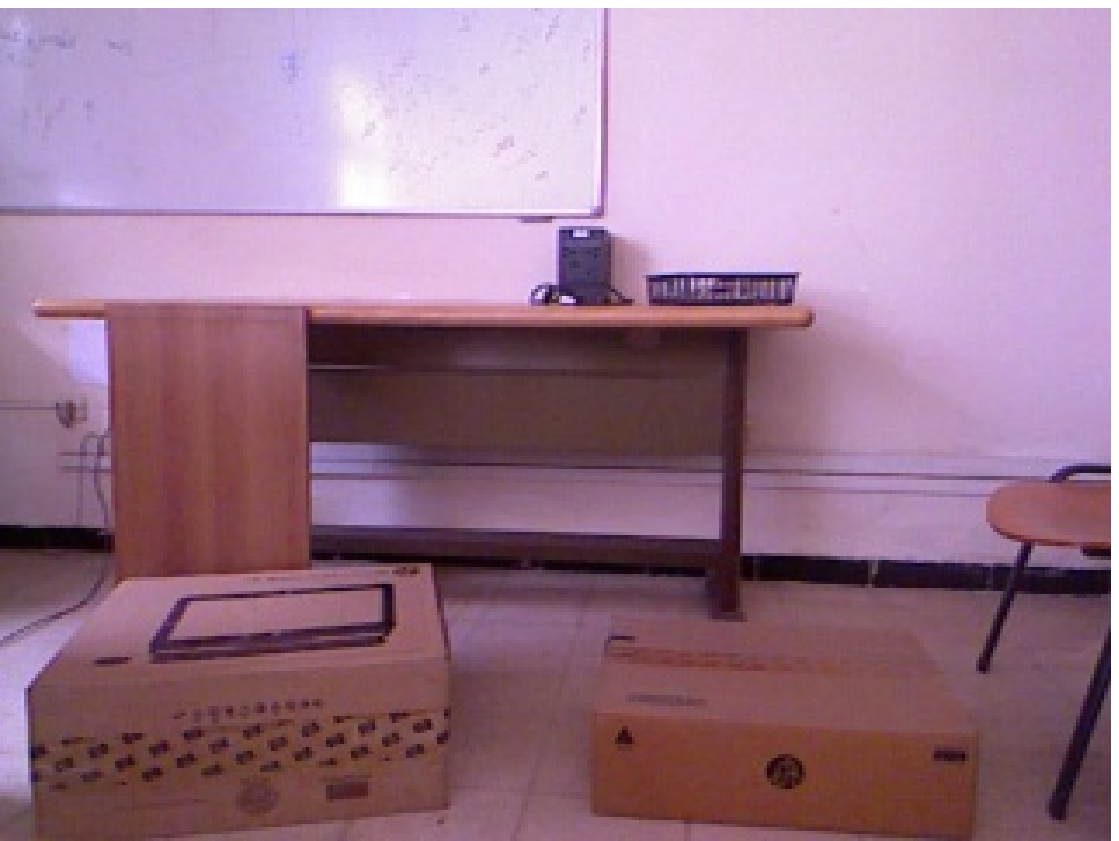}
		\includegraphics[height=2.5cm]{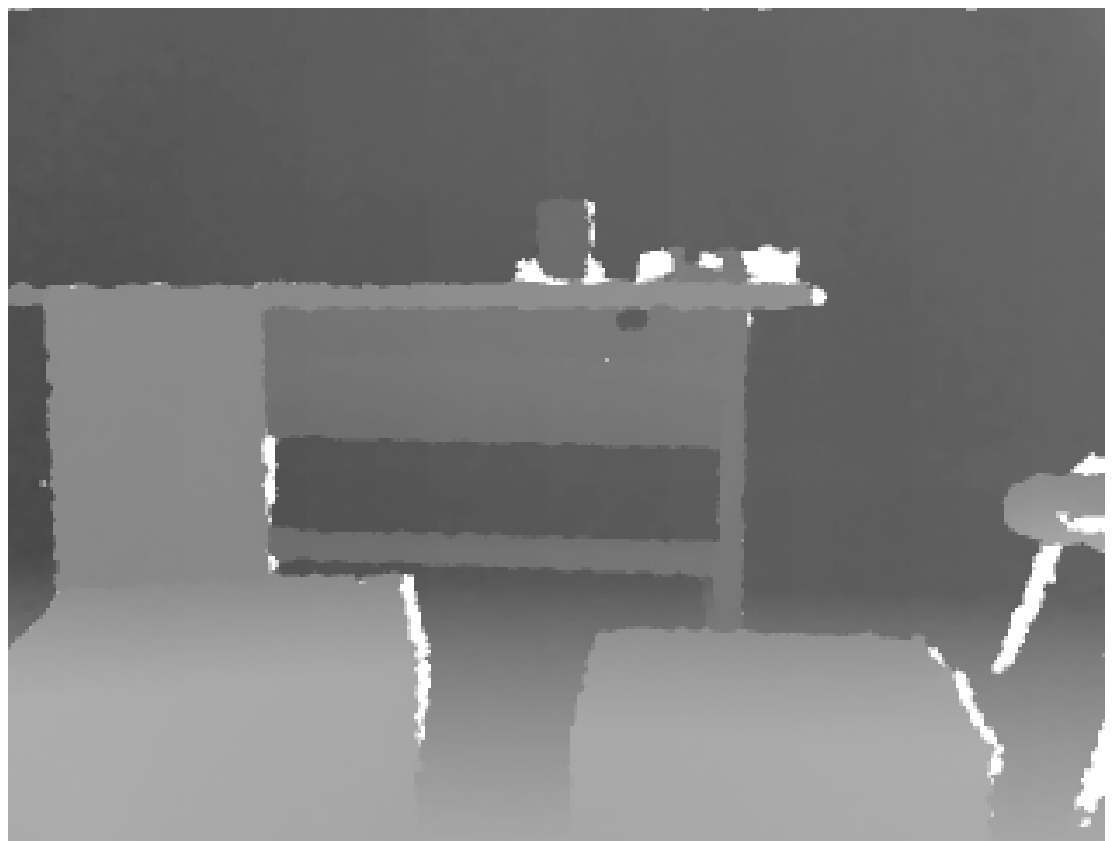}
		\includegraphics[height=2.5cm]{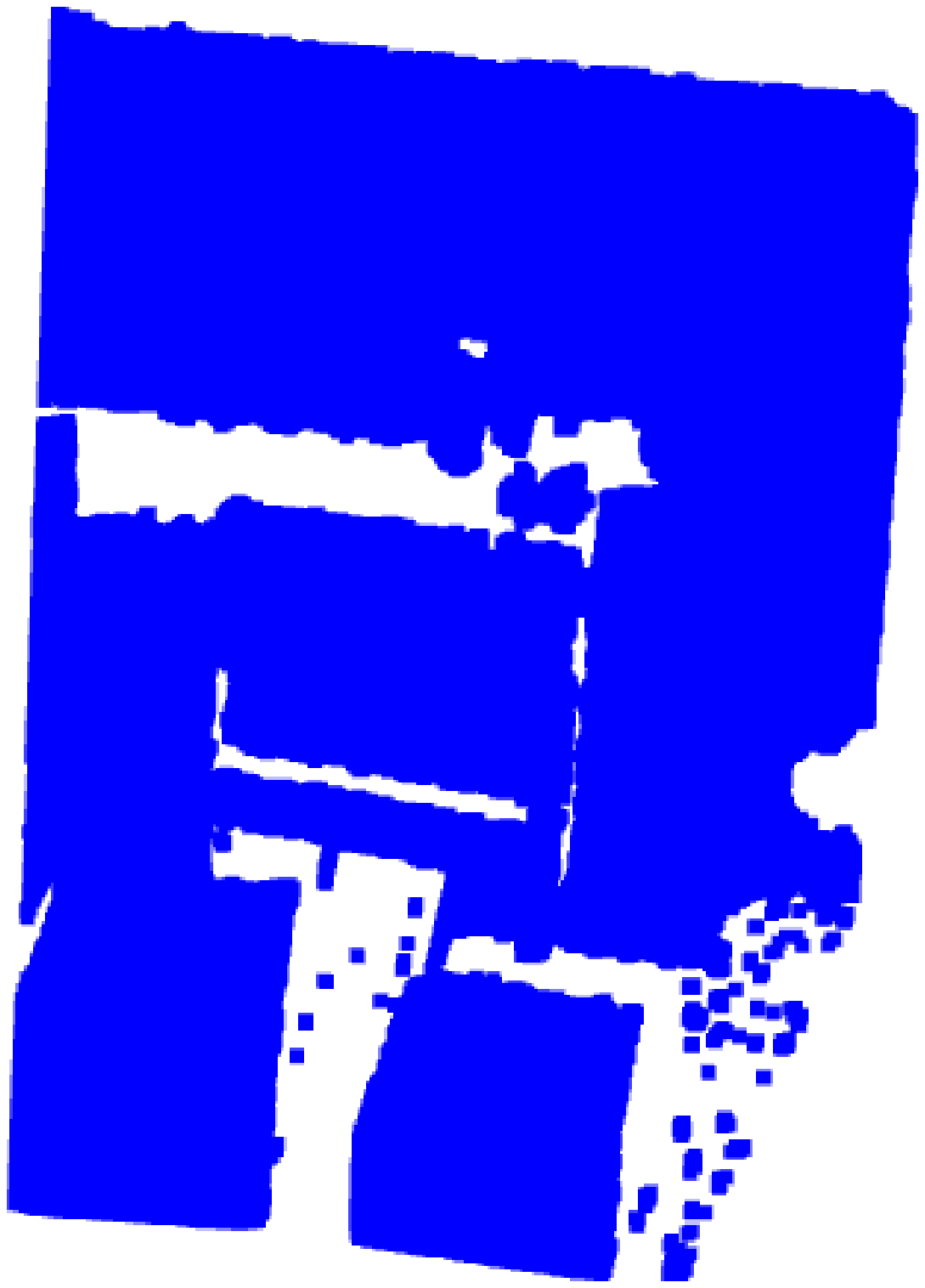}\\
		\includegraphics[height=2.5cm]{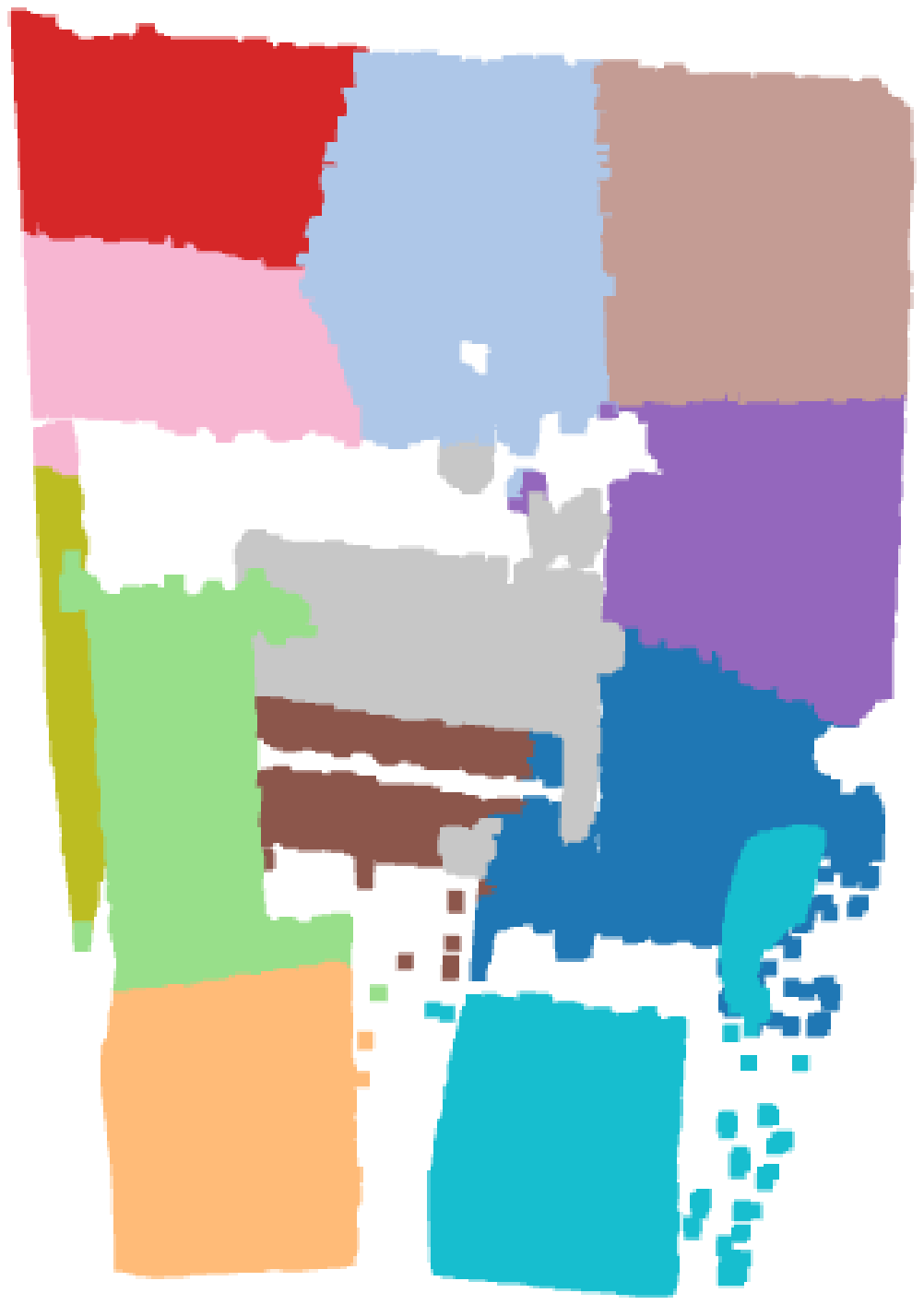}
		\includegraphics[height=2.5cm]{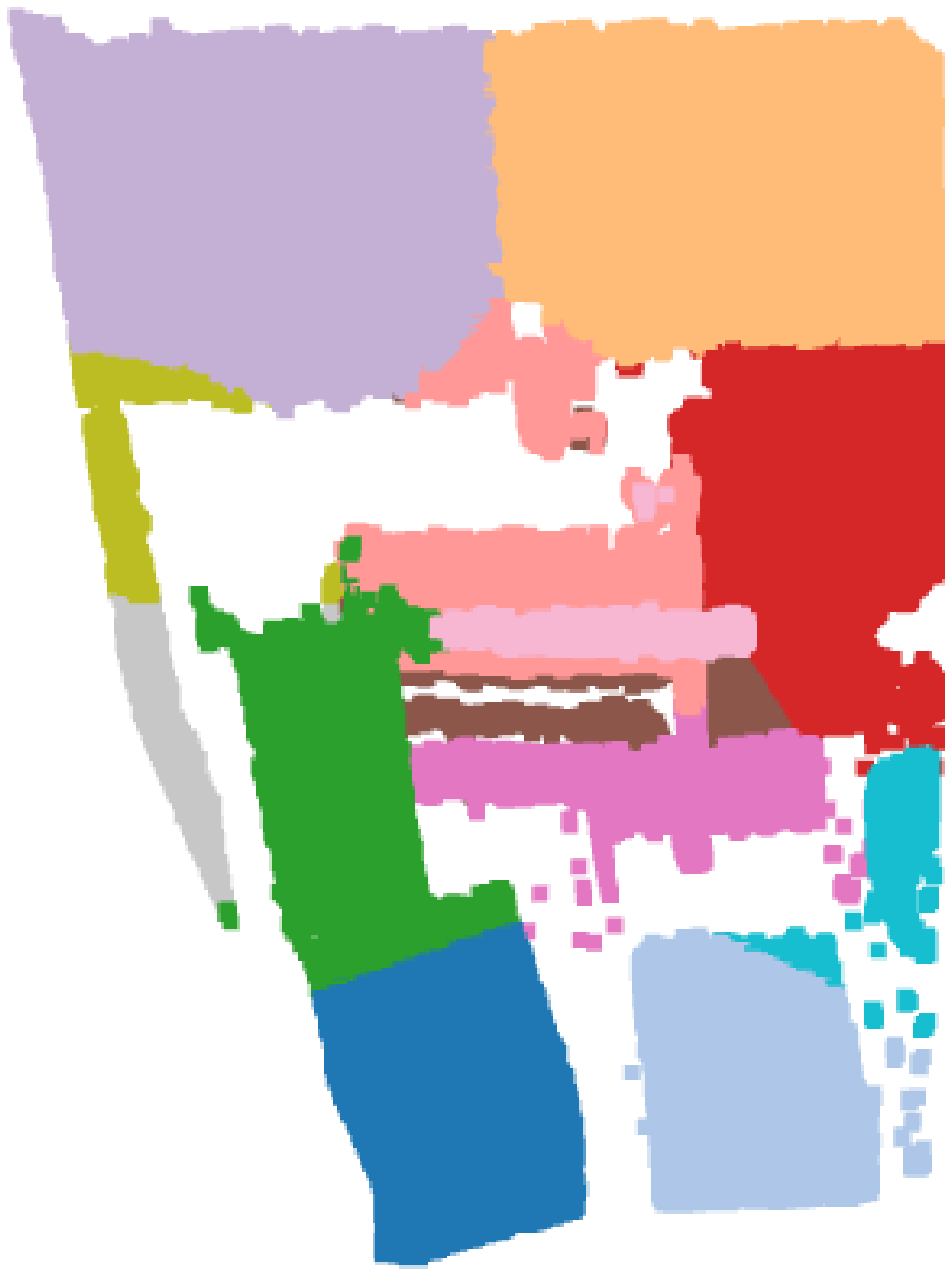}
		\includegraphics[height=2.5cm]{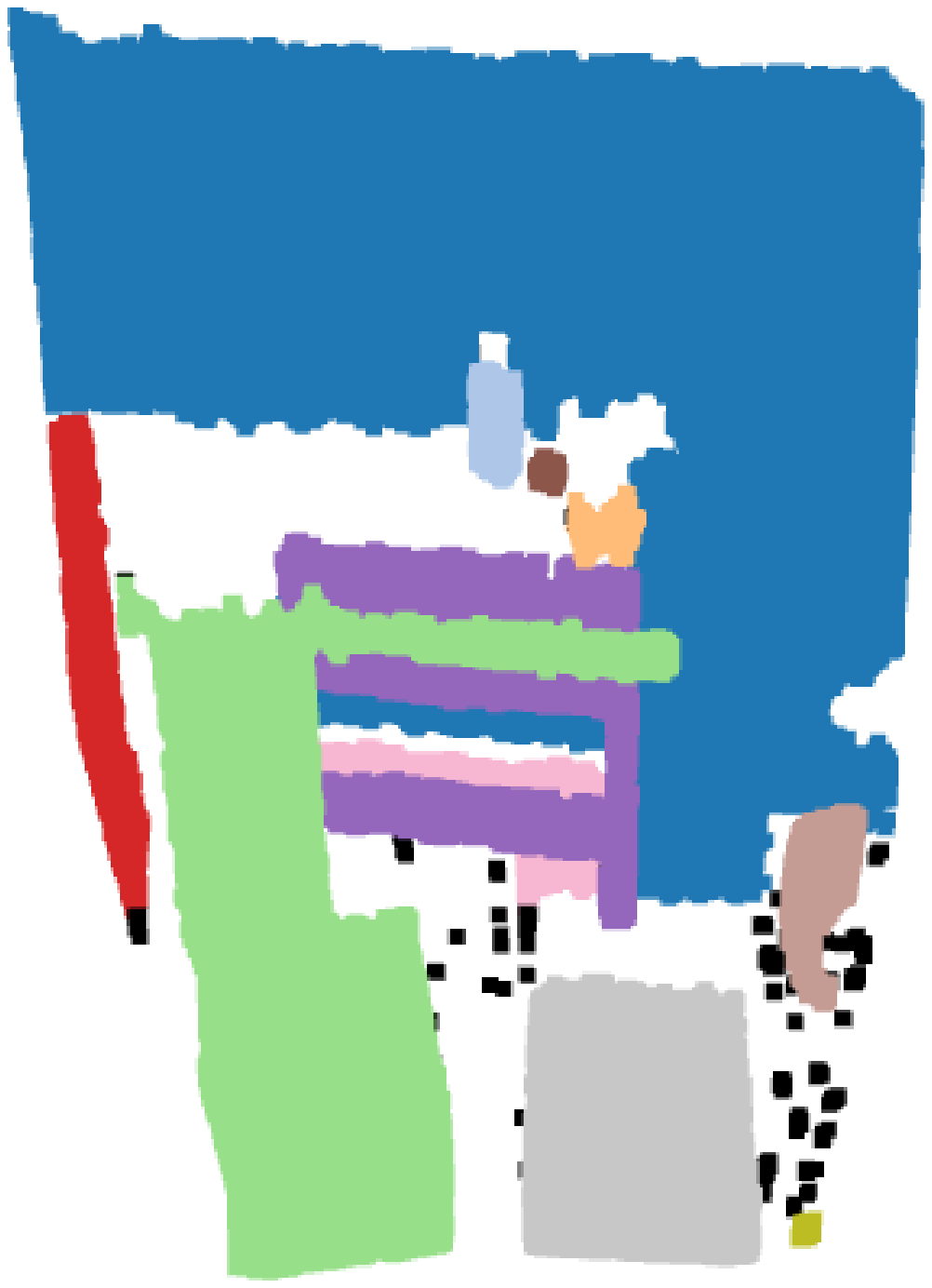}
		\caption{(Top) From left to right: color image, depth image and occupied space's point cloud. (Bottom) From left to right: K-means, Mean-shift and DBSCAN. The DBSCAN (the black points represent the detected noise).}
		\label{fig:segmentation}
	\end{figure}

In case of point cloud coarse segmentation, the DBSCAN is more suitable as shown by Figure \ref{fig:segmentation}. To reduce the temporal complexity, we applied a downsampling on the input point cloud. A pass-through filter can be applied for noise removal; its parameters are fixed according to the characteristics of the depth camera. In our case, we only considered points with a depth between $800mm$ and $4000mm$ apart. After the segmentation step, each segment is injected into the classification module and the feature extraction and classification module.
		
%%%%%%%%%% Conclusion:
In the next two sections, we will present in detail the main system components, namely the point cloud classification module, that is responsible for computing the geometric features and finding the point cloud nature using a deep neural network. Then, the semantic labeling module that is responsible for designing and synthesizing the scene using the objects classes that may serve as an output interface for assistive systems.
%%%%%%%%%%%%%%%%%%%%%%%%%%%%%%%%%%%%%%%%%%%%%%%%%%%%%%%%%%%%%%%%%%%%%%%%%%%%%%%%%%%%%%%%%%%%%%%%%%%%%%%
%						POINT CLOUD CLASSIFICATION
%%%%%%%%%%%%%%%%%%%%%%%%%%%%%%%%%%%%%%%%%%%%%%%%%%%%%%%%%%%%%%%%%%%%%%%%%%%%%%%%%%%%%%%%%%%%%%%%%%%%%%%
\section{Point cloud Classification in the context of assistive systems}\label{classification}
Our aim is to provide object classification in the context of assistive systems. In order to cover a significant number of classes in a simple way, we consider salient objects (large and medium objects) and regroup them into 7 semantic classes that include 16 object classes, namely chairs, beds, sofas, benches, stools, tables, desks, night-stands, dressers, wardrobes, shelves, bathtubs, toilets, windows, doors and stairs. These classes do not only contain semantic meaning, but the objects belonging to the same class have also a similar geometric shape while preserving some characteristics.
The first class represents objects that we sit on, it includes chairs, stools, beds, sofas and benches.
The second class represents objects that we put something on, it includes tables, night-stands and desks.
The third class represents objects that we put, hide or arrange something in, it includes dressers, wardrobes and shelves.
As for the fourth, the fifth and sixth classes, they represent bathtubs and toilets; windows and doors respectively. The last class represents stairs and since the stairs can be dangerous for the visually impaired and blind people, making the difference between the stairs leading to upstairs and stairs leading to downstairs is important.

In addition, this module includes 3 different sub-modules. The first sub-module computes the point cloud semantic class based on classifying its global features computed using a deep neural network. The second sub-module computes the point cloud geometric features. As for the last sub-module, it combines the output of these sub-modules in order to provide the final classes and provide them for the semantic labeling module.
\subsection{Semantic classification}
We call the semantic classification/classes, the classification/classes provided by the deep neural network in order to distinguish between this later and the final classification that is mapped on the scene synthesis. 

Deep learning architectures showed considerable improvement in many fields, especially with RGB data after introducing the Convolutional Neural Networks, CNNs. Current works investigate the 3D object classification using volumetric data \cite{kumawat2019lp}\cite{maturana2015voxnet}\cite{wu20153d} or point cloud \cite{liu2019relation}\cite{li2018so}\cite{qi2017pointnet++}\cite{qi2017pointnet} as the input. The advantage of using point clouds based networks instead of CNNs based or volumetric based networks is the fact that point clouds are irregular, most of the time sparse and permutable.

Note that, in this work, we are not designing a novel architecture. However, using our approach of merging the objects classes that are geometrically similar and have the same function increased the accuracy with 5\% on the test set. The obtained results are reported, compared and discussed in the experiments section. To classify a given point cloud, we trained the Pointnet network \cite{qi2017pointnet}, a state of the art architecture. Pointnet is a deep multi-layer perceptron network that is designed for 3D raw data. It is invariant to permutation and transformation due to the use of max pooling, a symmetric activation function, and T-net that ensures pose normalization. It is also robust to small noise and incomplete data (with a small portion), but it is not able to capture fine local patterns. This limitation will not highly affect our system since we only consider salight objects (for the time being). Although several architectures have been proposed over time such as \cite{liu2019relation} \cite{li2018so}, pointnet is still used in different systems that involve point clouds \cite{ramasinghe2019spectral} \cite{milz2019points2pix}.
\subsection{Geometric features extraction}
In addition to the discussed object classification, we extract from the computed convex hull geometric features, namely object's height and its occupied area as explained in \cite{zatout2019ego}. This is done in the online mode, the offline mode is not needed for this classification. After the segmentation process, extracting geometric features and the classification is computed in parallel with the object classification. This object description is useful for the visually impaired and blind people, it helps them to have an impression about their surroundings.
\subsection{Point cloud classification}\label{pccls}
We combined these two classification methods, i.e. the classification using deep learning and the classification using the geometric features to provide rich information. For each object, we provide its semantic class, its height class and its occupied area class. When the deep learning model predicts the object class with low probability, only the second classification is maintained. In this way, we are sure to provide an accurate description even if the deep learning model fails to predict the right semantic class.

In addition, combining these classes allows the distinction between objects belonging to the same semantic class. Chairs, beds, sofas and benches are objects that we sit on; however, they are different in their forms and occupied area: beds are generally a large object (3rd class regarding the occupied area), chairs have small surface (1st class regarding the occupied area) and benches and sofas are medium objects (2nd class regarding the occupied area).
Dressers, night-stands and shelves are all objects that we arrange something in; however, they have different heights and occupied areas: dressers are represented by the 3rd class regarding the height, night-stands are represented by the 2nd class, but shelves are usually medium objects and night-stands are usually small.
%%%%%%%%%%%%%%%%%%%%%%%%%%%%%%%%%%%%%%%%%%%%%%%%%%%%%%%%%%%%%%%%%%%%%%%%%%%%%%%%%%%%%%%%%%%%%%%%%%%%%%%
%						SEMANTIC LABELING
%%%%%%%%%%%%%%%%%%%%%%%%%%%%%%%%%%%%%%%%%%%%%%%%%%%%%%%%%%%%%%%%%%%%%%%%%%%%%%%%%%%%%%%%%%%%%%%%%%%%%%%
\section{Object semantic labeling and scene synthesis}\label{Semantic}
In this section, we first introduce our proposed semantic labeling that represents the class of each detected object. Then, we describe how the perceived 3D scene is synthesized into a specific area while taking into account the located ground, the semantic label and extracted features of each 3D object.
\subsection{Semantic labeling}
In our previous work \cite{zatout2019ego}, we proposed our first semantic labeling: each horizontal plane in the scene was represented as cylinders with an associated height and radius that allow representing the occupied area and the height of the associated object respectively.
This allows the user to have an impression about his surroundings: free space, small objects, medium objects, large objects and their heights, but it does not provide the object's shape nor its nature: if it is a chair, a table, etc.

\begin{figure}[!t]
	\centering
	\includegraphics[width=4.75cm]{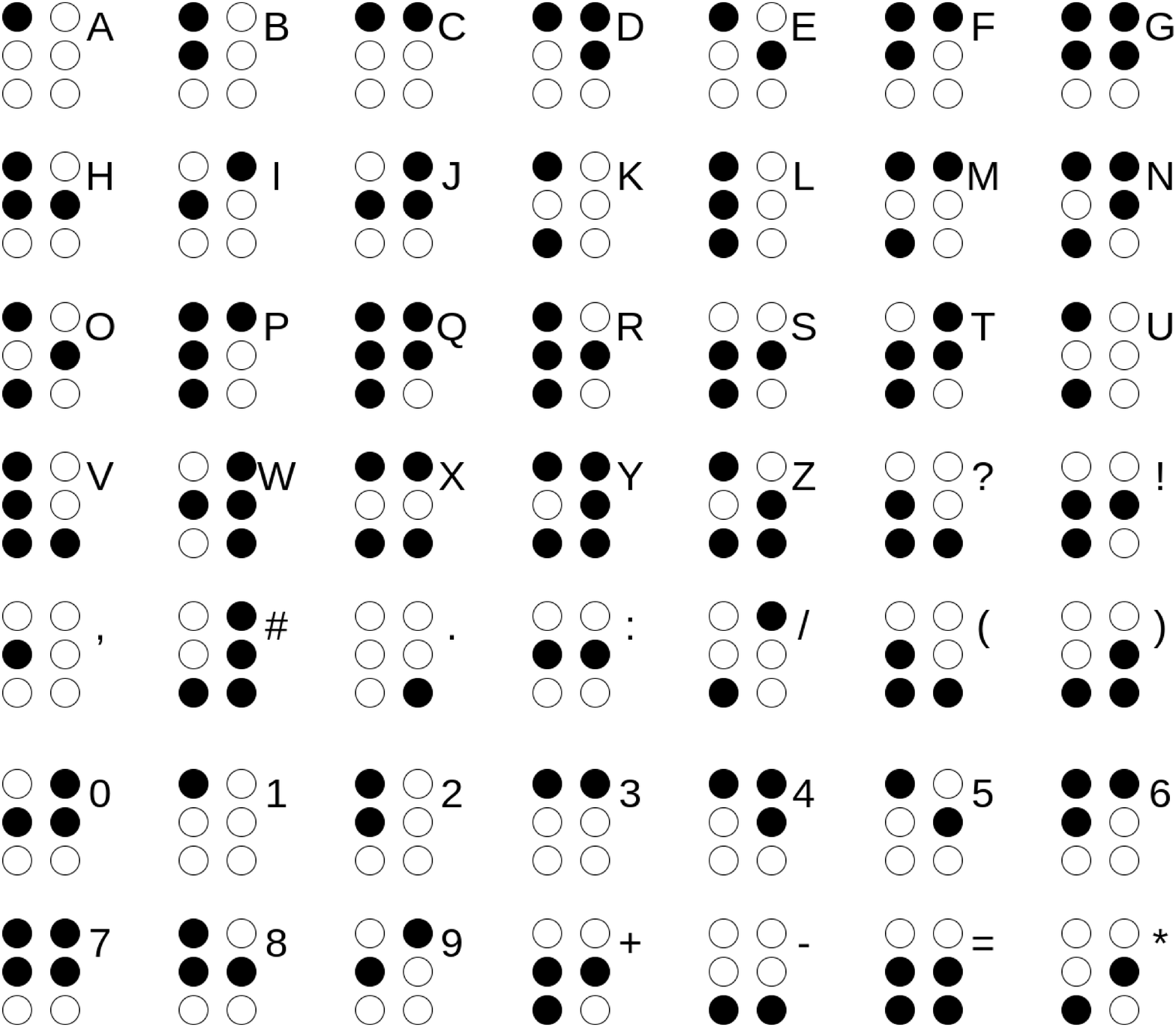}
    \includegraphics[width=5.25cm]{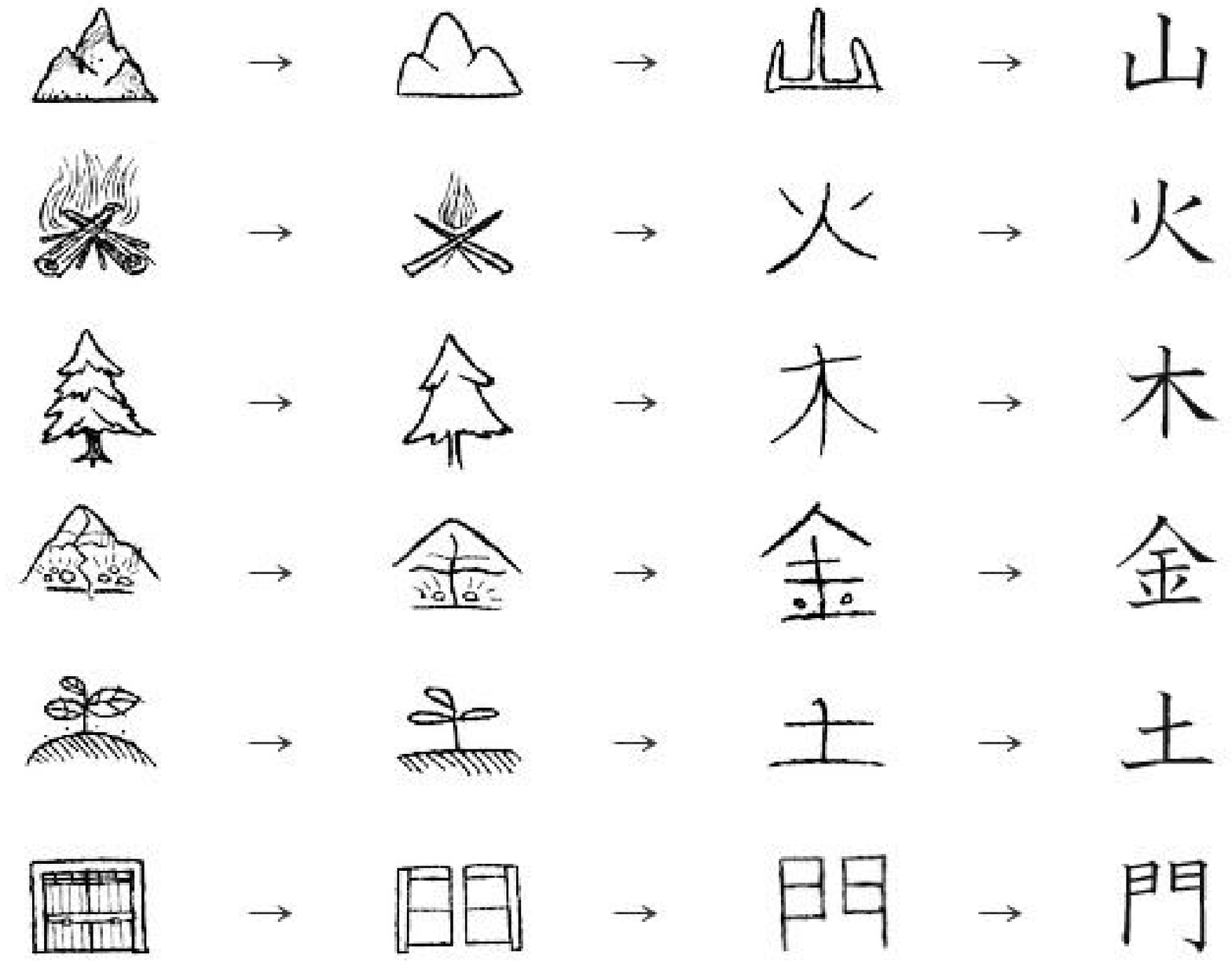}
    \caption{(Left) The Braille system. (Right) Kanji, the Japanese writing system.}
	\label{fig:braille_kanji}
\end{figure}

In this work, we propose an improved semantic labeling which takes into consideration the nature of the objects. It is inspired by the \textbf{Braille system} and \textbf{Kanji} (the Japanese writing system)(Figure \ref{fig:braille_kanji}). An alphabet in the Braille system is represented by a cell that is provided with raised dots. Each cell contains at most six raised dots. \textbf{Kanji} is a Japanese writing system that is inspired by logographic Chinese characters. Some Kanji letters that represent some objects are driven from nature, i.e. these objects' shape in real-world; it's the case for 'mountain' as shown in Figure \ref{fig:braille_kanji} (right) first line.

In order to draw our illustrative semantic labeling, we designed cells with at most 25 raised dots. The cell's shape can be revised depending on the precision of the synthesis area: if the synthesis area is rich in pins, we can design cells with more raised dots. We suggest to use cells such that the shape can be touched only by a single finger to avoid ambiguity.

\begin{figure}[!t]
	\centering
	\begin{tabular}{c|c}
	\includegraphics[width=4cm]{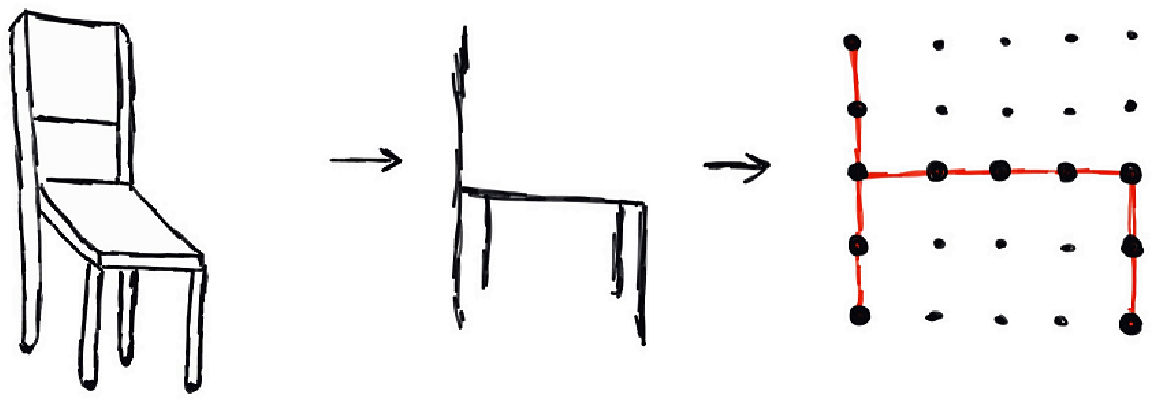}&
	\includegraphics[width=4cm]{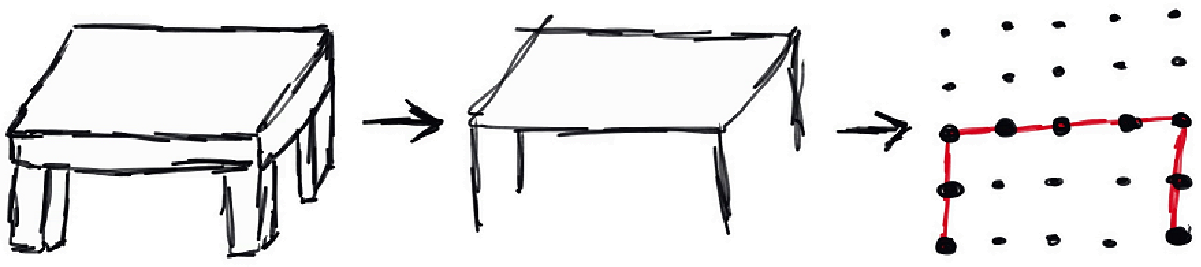}\\
	\includegraphics[width=4cm]{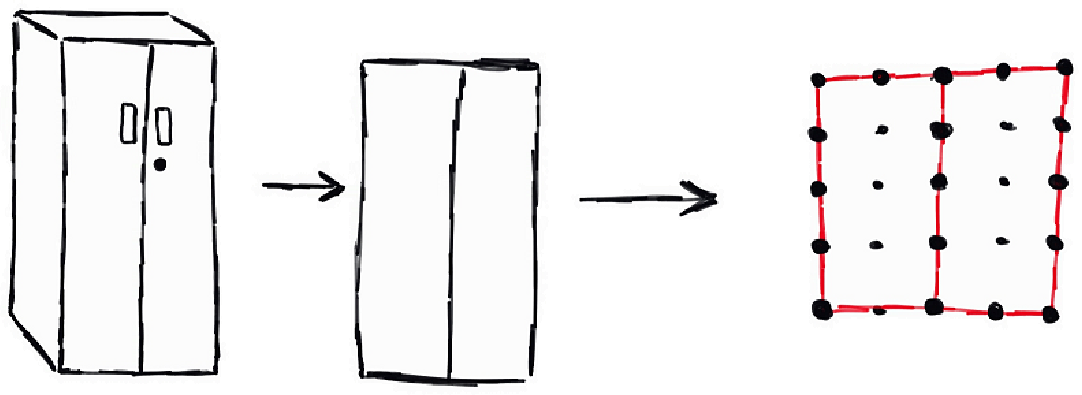}&
	\includegraphics[width=4cm]{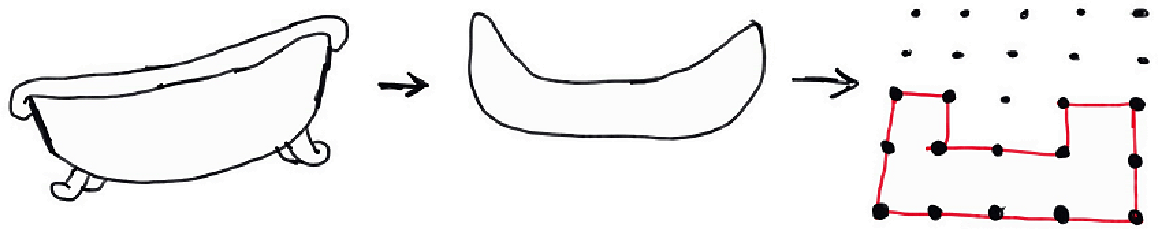}\\
	\includegraphics[width=4cm]{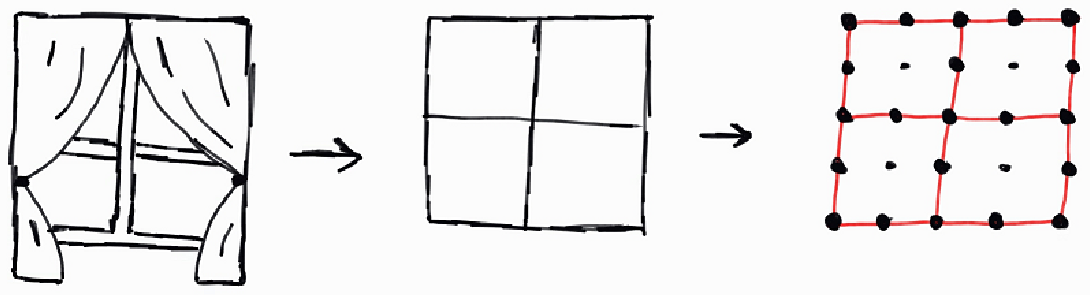}&
	\includegraphics[width=4cm]{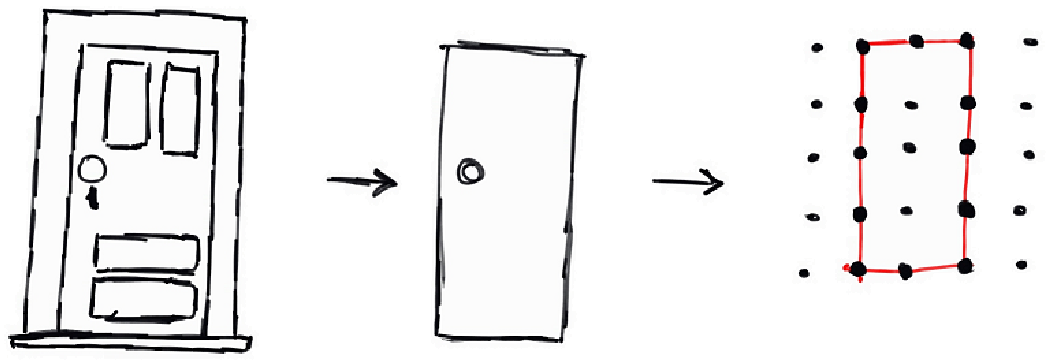}
	\end{tabular}
    \caption{Our proposed semantic labeling from the first class to the last class, respectively (From top to down). We first drew the selected objects in the real-world (Left) and then, we derived shapes recursively until we obtained the current semantic labeling (From left to right).}
	\label{fig:semanticlabeling}
\end{figure}

\begin{figure}[!t]
	\centering
	\begin{tabular}{c|c}
	\includegraphics[width=4cm]{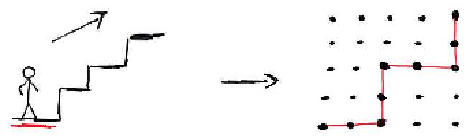}&
	\includegraphics[width=4cm]{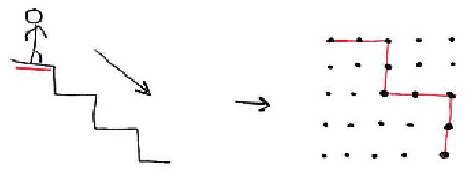}
	\end{tabular}
    \caption{The proposed semantic labeling for stairs. Stairs leading to upstairs and leading to downstairs, their chosen labels, their positioning on the synthesis area.}
	\label{fig:stairs}
\end{figure}

On the other hand, the illustrative labeling is derived from the object's shape in the real-world. For each semantic class, we chose the object that is the most close to the class' meaning: we chose a chair, a table, a dresser, a bathtub, a window and a door to represent the proposed classes in section \ref{classification} respectively as shown in Figure \ref{fig:semanticlabeling}. To ensure the user's safety, we provide two different labels for the stairs leading to upstairs and stairs leading to downstairs as shown in Figure \ref{fig:stairs}. In this way, by touch, the user will understand if he is about to climb the stairs or about to go downstairs.

The described labeling is used alongside with the first labeling to enrich the scene description. The reason we combined the two labels instead of creating an illustrative label for each object, is to reduce the number of possible illustrative labels since it can be deducted from its height and its occupied area as explained in section \ref{pccls}. Note that although the labels are inspired by Braille and Kanji systems, understanding and memorizing these labels do not require the user to be comfortable with them.
\subsection{Scene synthesis}
After detecting the ground and classifying the computed segments, the system provides the ground pixels and a set of segments. Each segment is associated with its label and its extracted geometric features. These latter will be used for scene synthesis.

The area of the scene synthesis has a trapezoid shape that corresponds to the scaled view field of the camera. This area is covered by a dense number of pins that represent the raised dots in the Braille system.
The pin  is set up to \textbf{5} different levels: The level zero is used to represent the holes on the ground. The level 1 represents the ground or the neutral element in some applications. The height of the illustrative labels that are mapped in the center of the convex hull, is set to the computed level (i.e. level 2, 3 or level 4) and the height of the remaining area of the convex hulls associated with the segment is set to level 2.

\begin{figure}
	\centering
	\includegraphics[width=8cm]{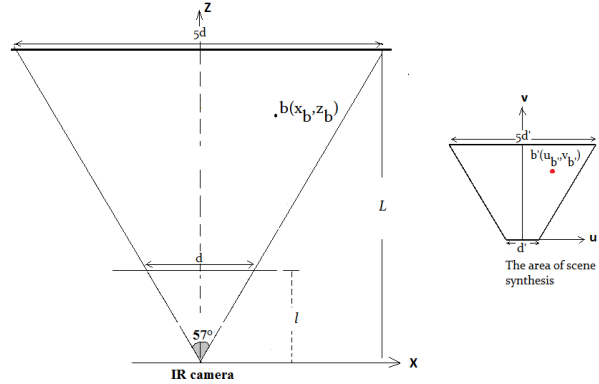}
		\caption{(Left) The depth camera's view field. (Right) The area of scene synthesis where semantic labels are generated.}
		\label{fig:devicep}
	\end{figure}

The parameters of the depth camera are used to map the view field into an area of scene synthesis (see Figure \ref{fig:devicep}). Knowing that the ratio between the small and great basis of the trapezoid representing the camera's view field is equal to $5$ (for $L=400cm$), the scaled area of scene synthesis verifies the same ratio as indicated by Figure \ref{fig:devicep})(right).

Let $b(x_b,y_b,z_b)$ be the barycenter of the box encompassing the object. The position of $b'(u_{b'},v_{b'})$, the mapping of $b$ on the synthesis area is calculated from the geometric relationship between the scene and the synthesis area as indicated by Figure \ref{fig:devicep}. The values of $u_{b'}$ is equal to $d' \times x_b/d$ and $v_{b'}$ is equal to $d' \times (z_b-l)/d$ where $d'$ is the small basis of the synthesis area. The values of $l, L$ for the Kinect sensor are $80cm$ and $400cm$.

Now, let $b_1(x, y), b_2(x, y), .., b_n(x, y)$ be the points defining the box encompassing the object. The mapped points $b'_1(u, v), b'_2(u, v), .., b'_n(u, v)$ are located on the area. All pins inside the area defined by the points $b'_i, i=1..n$ are set to level $2$. The pins associated to the label of the object are set to their associated level $3$, $4$ or $5$ at the barycenter as indicated by Figure 8 (Left). 

\begin{figure}[!t]
	\centering
	\includegraphics[height=2.5cm]{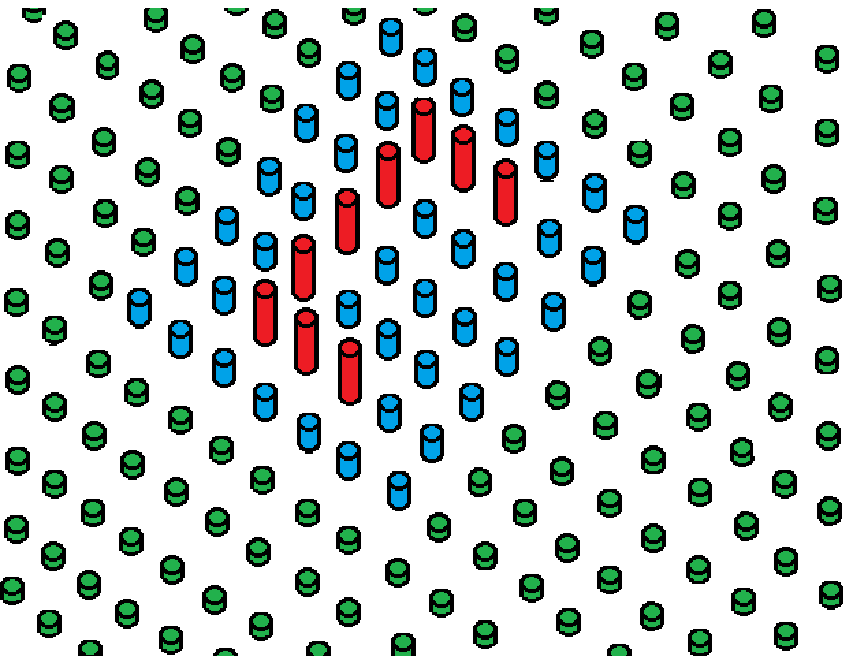}
	\includegraphics[height=2.5cm]{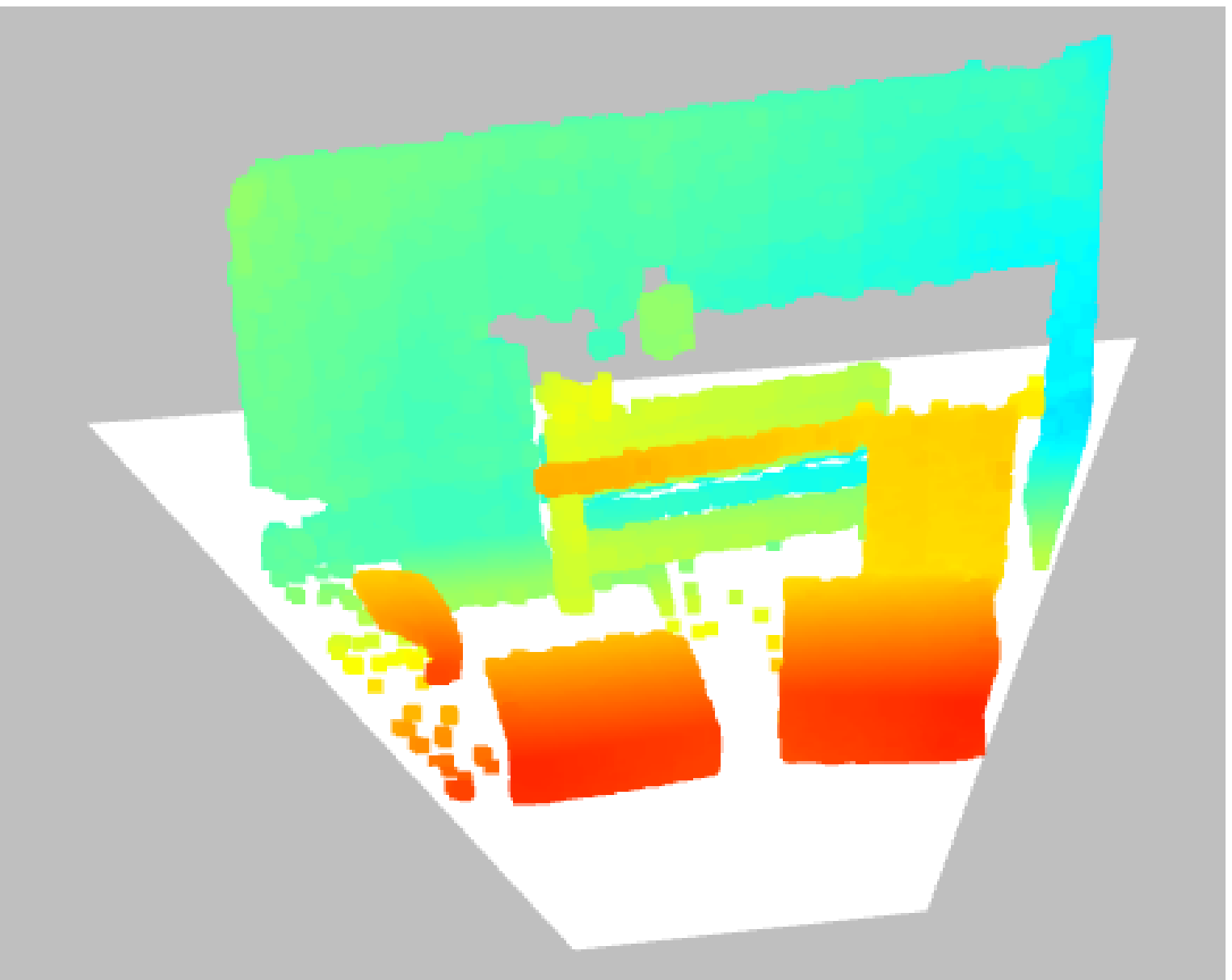}
	\includegraphics[height=2.5cm]{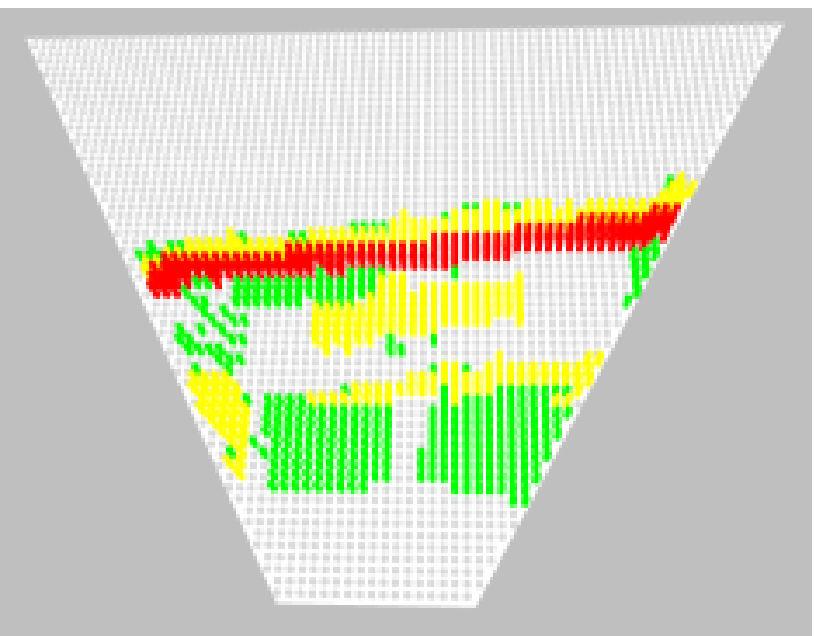}
\caption{(Left) The convex hull and the positioning of the label (here for a table) on the area. Note that the pins of the label in red are of level 3 (corresponding to the height of the table in the scene). The pins of the convex area in blue are of level 2. The remaining pins in the synthesis area (in green) are set to level 1 corresponding to the ground. (Middle) Mapping the 3D point cloud into the synthesis area. (Right) Scene synthesis from raw data (point cloud) using pins without semantic labeling.}
	\label{fig:bassar1}
\end{figure}

The area's input may receive the 3D raw data or labels. In Figure \ref{fig:bassar1} (Middle), the point cloud of the captured scene is directly mapped on the area. The same point could be mapped into the synthesis area using different heights of pins (see Figure \ref{fig:bassar1} (Right)). For more visibility, we assigned the grey color, the green color, the red yellow color and the red color to represent objects with level 2 (ground), 3, 4 and 5 respectively.
%%%%%%%%%%%%%%%%%%%%%%%%%%%%%%%%%%%%%%%%%%%%%%%%%%%%%%%%%%%%%%%%%%%%%%%%%%%%%%%%%%%%%%%%%%%%%%%%%%%%%%%
%						EXPERIMENTS
%%%%%%%%%%%%%%%%%%%%%%%%%%%%%%%%%%%%%%%%%%%%%%%%%%%%%%%%%%%%%%%%%%%%%%%%%%%%%%%%%%%%%%%%%%%%%%%%%%%%%%%
\section{Experiments}\label{exp}
\subsection{Experimental environment and dataset}
In order to perform the training, we ran our models on a GPU provided by the Google Colaboratory platform. After that, we executed our proposed system on a laptop having Intel Core $i5-7200U$ CPU $(2.50GHz * 4)$ and $4GB$ as processor and RAM respectively.

\begin{figure}[t]
	\centering
		\begin{tabular}{c}
		\includegraphics[height=2.5cm]{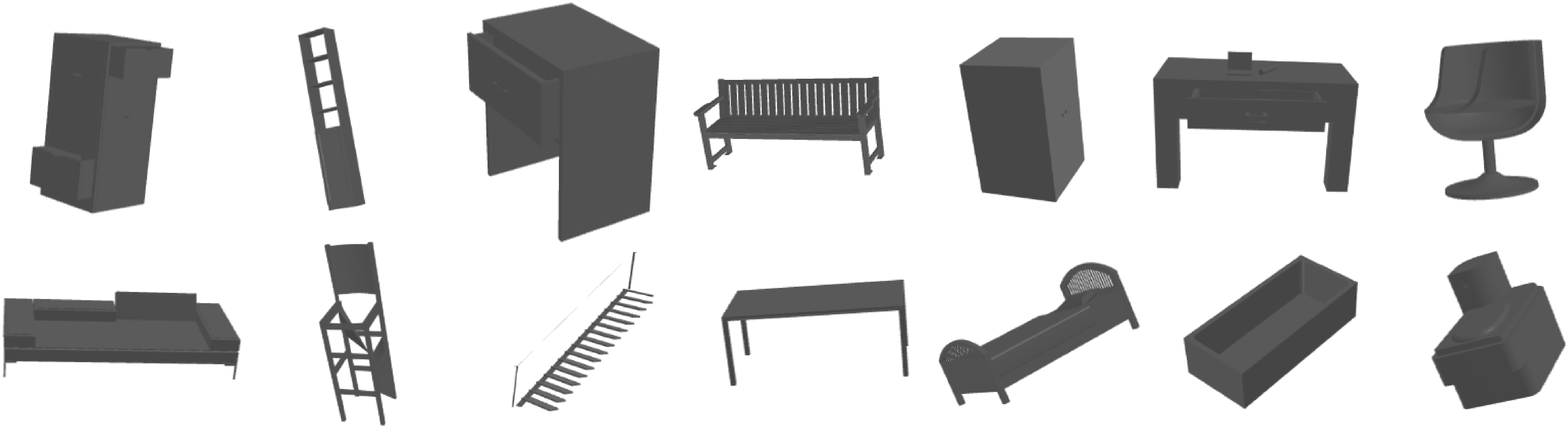}\\
		\hline
		\includegraphics[width=9cm]{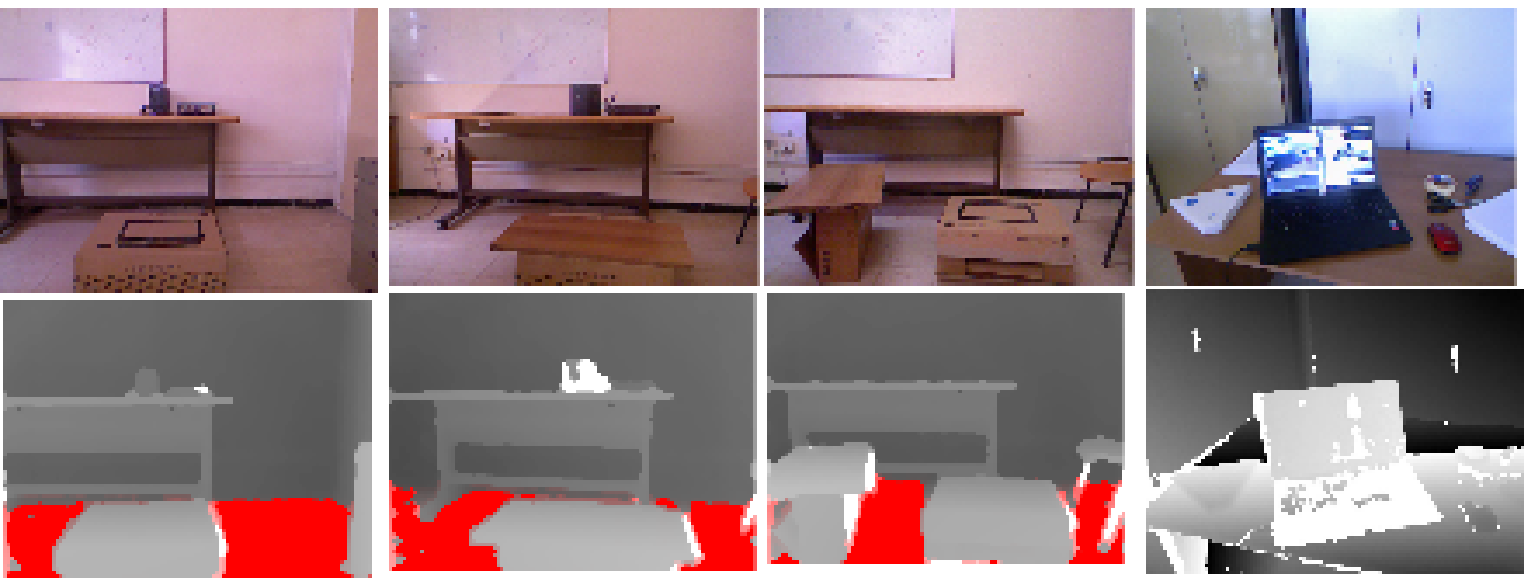}
		\end{tabular}
		\caption{Dataset samples. (TOP) Modelnet40\cite{wu20153d}. (Down) GDIS\cite{GDIS2019}: the red color represents the detected ground.}
		\label{fig:modelnet40}
\end{figure}

As to train our model, we extracted our 16 selected classes from ModelNet40 \cite{wu20153d} (Figure \ref{fig:modelnet40}), a public dataset which consists of 40 classes of CAD models. Note that we excluded doors and windows since they are represented by holes and in the real-world can be confused with noise or walls where they are closed. ModelNet40 originally includes 12311 models among which 2468 are used for testing. Our constructed  dataset includes 5560 models among 988 which are used for testing.

In addition, we used GDIS dataset (Ground Detection for Indoor Scenes) \cite{GDIS2019}, our local dataset, for additional experiments. GDIS dataset was initially constructed for the ground detection task, but we considered some scenes for point cloud classification and the validation of the entire system (from the ground detection to the semantic labeling).
\subsection{Point cloud segmentation and feature extraction}
Since we consider irregular segments and not planes, we computed the height as the height of the 90th percentile of the points' $y-coordinate$. We chose the 90th percentile instead of the 100th to avoid potential outliers. To evaluate this step, we took the measurements of some objects in the real-world and then compare it with the obtained geometric features. This is done by computing the Mean Absolute Error (MAE) between the objects' height and the computed one. The obtained results showed a slight mean difference that does not exceed $30mm$.
\subsection{Semantic classification}
\begin{figure}[t]
	\centering
    \includegraphics[width=5cm]{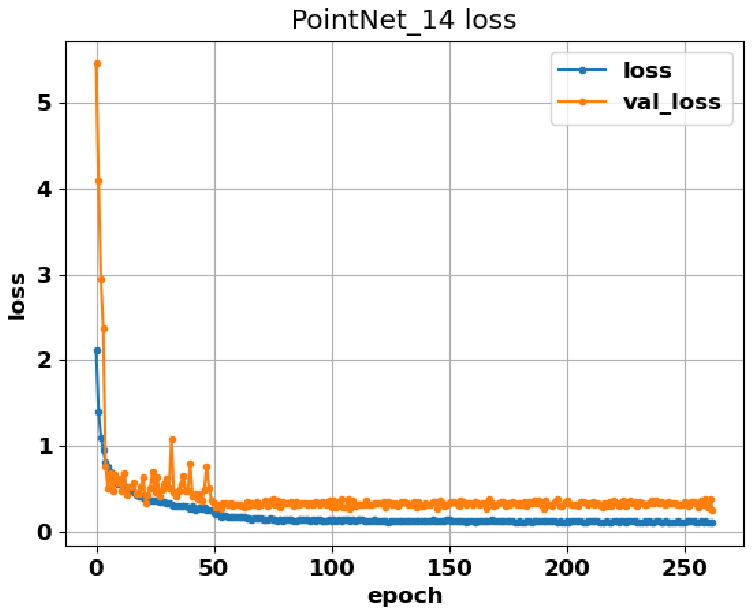}
    \includegraphics[width=5cm]{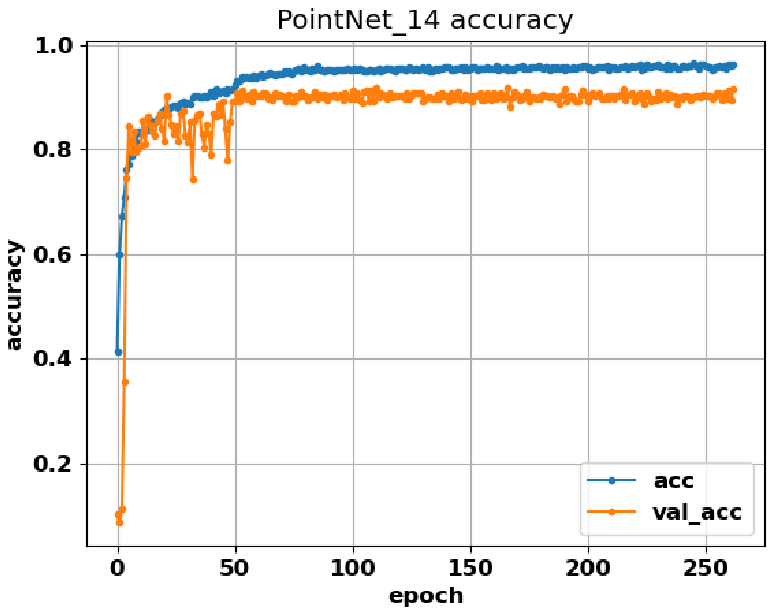}
    \caption{(Left) PointNet loss function. (Right) PointNet accuracy. Training PointNet with 14 classes; the model starts to converge from epochs 50 and reach 96.19\% and 91.67\% on train-set and test-set respectively.}
		\label{fig:pointnet_14}
\end{figure}

% compare loss functions and accuracy 
Since PointNet is 3D raw data based, the CAD models are sampled into point clouds of 2048 points and normalized into a unit sphere. Regarding the experimental settings, we applied the same configurations as in \cite{qi2017pointnet}. As for data preprocessing, we, on-the-fly, applied random rotations around the yaw axis and jittered  the points position as described in \cite{qi2017pointnet}. PointNet achieved the state-of-the-art with 89.2\% accuracy in their original paper. In this work, we trained our model on a sub-set of 14 (after excluding doors and windows) classes from ModelNet40. The model started to converge from the 50th pechos and reached 91.67\% accuracy on the test-set after 262 epochs (Figure \ref{fig:pointnet_14}).

\begin{figure}[t]
	\centering
    \includegraphics[width=10cm]{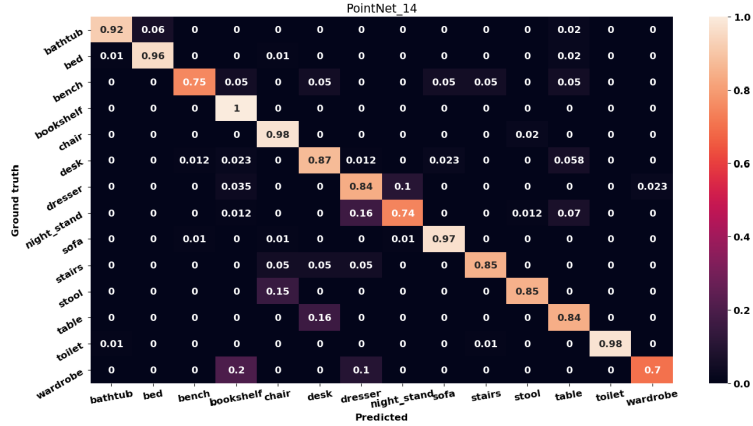}
    \caption{Confusion matrix: there is high confusion between wardrobe and the classes bookshelf and dresser (about 20\% and 10\% respectively), between table and desk (about 16\%), stool and chairs (about 15\%); and small confusion between the sofa and the classes bench and chair.}
	\label{fig:confmat_pnet_14}
\end{figure}

\begin{figure}[t]
	\centering  
    \includegraphics[width=5cm]{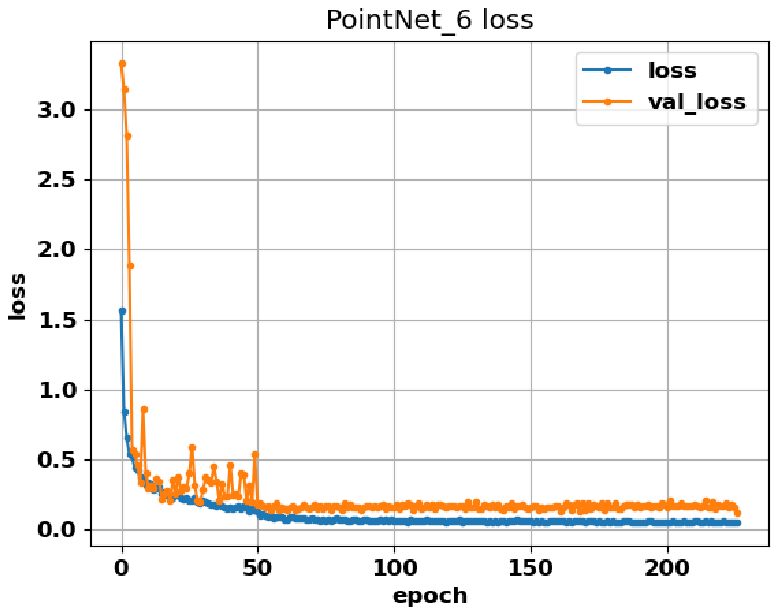}
    \includegraphics[width=5cm]{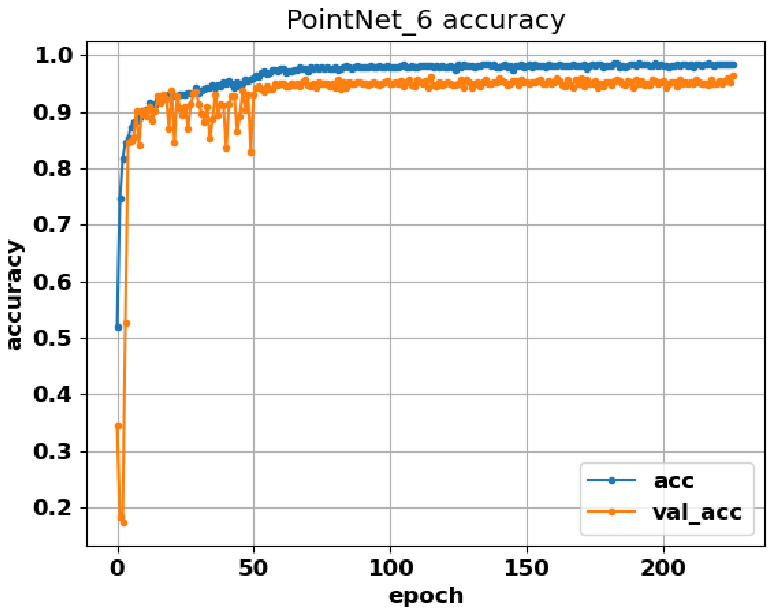}
    \caption{(Left) PointNet loss function. (Right) PointNet accuracy. Training PointNet with 6 classes; the model starts to converge from epochs 50 and reach 98.26\% and 96.35\% on train-set and test-set respectively.}
		\label{fig:pointnet_6}
\end{figure}

\begin{figure}[t]
	\centering
    \includegraphics[width=5cm]{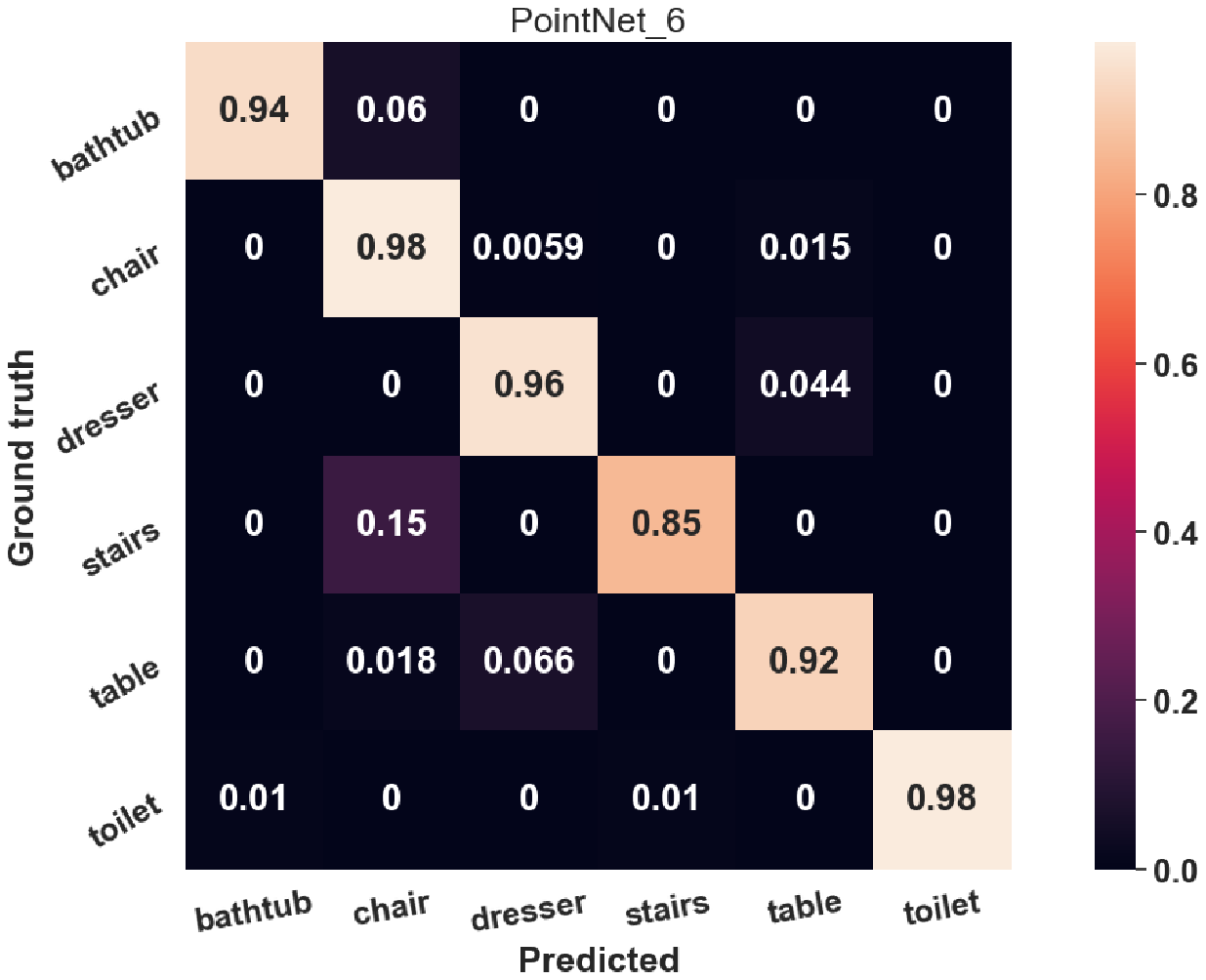}
    \includegraphics[width=5cm]{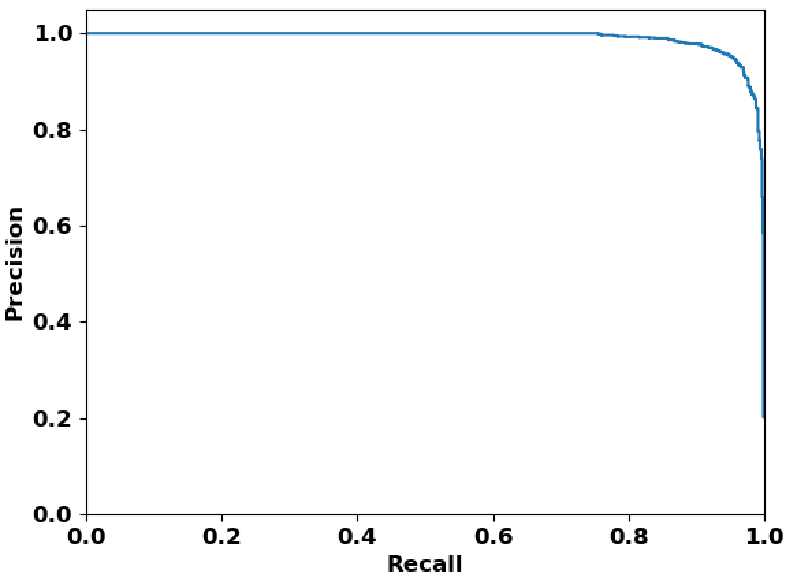}
    \caption{(Left) Confusion matrix: the confusion between objects does not exceed 6\% unless for stairs that 15\% of them were confused as chairs. Most of the objects are not confused with other objects (0\% confusion). (Right) Recall-precision curve: the system predicts accurately with high recall and high precision).}
		\label{fig:confmat_pnet_6}
\end{figure}

\begin{figure}[t]
	\centering
    \includegraphics[width=9cm]{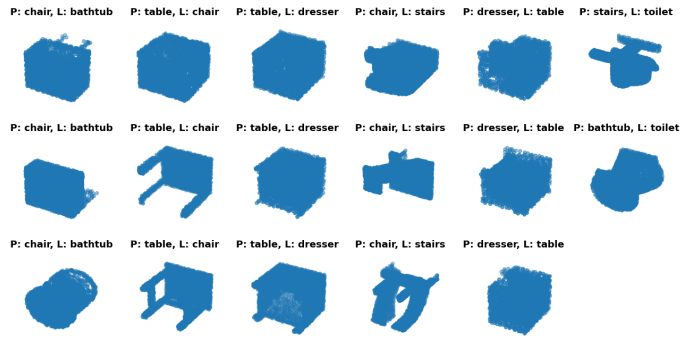}
    \caption{Worst 3 classification results. P and L stand for Predicted and Label respectively. There is a geometric resemblance of the input instance with the objects of the predicted class.}
		\label{fig:worst}
\end{figure}

However, as shown by the confusion matrix (see Figure \ref{fig:confmat_pnet_14}), there is a high confusion between wardrobe and the classes bookshelf and dresser (about 20\% and 10\% respectively), between table and desk (about 16\%), stool and chairs (about 15\%). There is also a small confusion between the sofa and the classes bench and chair. These confusions are due to the geometric resemblance between them. This leads us to consider combining classes for the semantic classification as explained earlier in section \ref{classification}. As the previous model, this model started to converge after the 50th epoch. The accuracy has been improved by 5\% to reach 96.35\% on the test-set after only 226 epochs (Figure \ref{fig:pointnet_6}). Thus, the described confusions have been eliminated (see Figure \ref{fig:confmat_pnet_6}). The confusion matrix (see Figure \ref{fig:confmat_pnet_6} (Left)) shows that the confusion between the different classes does not exceed 7\% unless for stairs that 15\% of them were confused as chairs. To visualize this latter, we plot the three worst classification for each class (Figure \ref{fig:worst}): the instances have similar geometry with the predicted class with a certain probability. However, the confusion stills small and the system predicts accurately with high recall and precision (see Figure \ref{fig:confmat_pnet_6} (Right)). 

\begin{figure}[!t]
	\centering
\includegraphics[width=2cm]{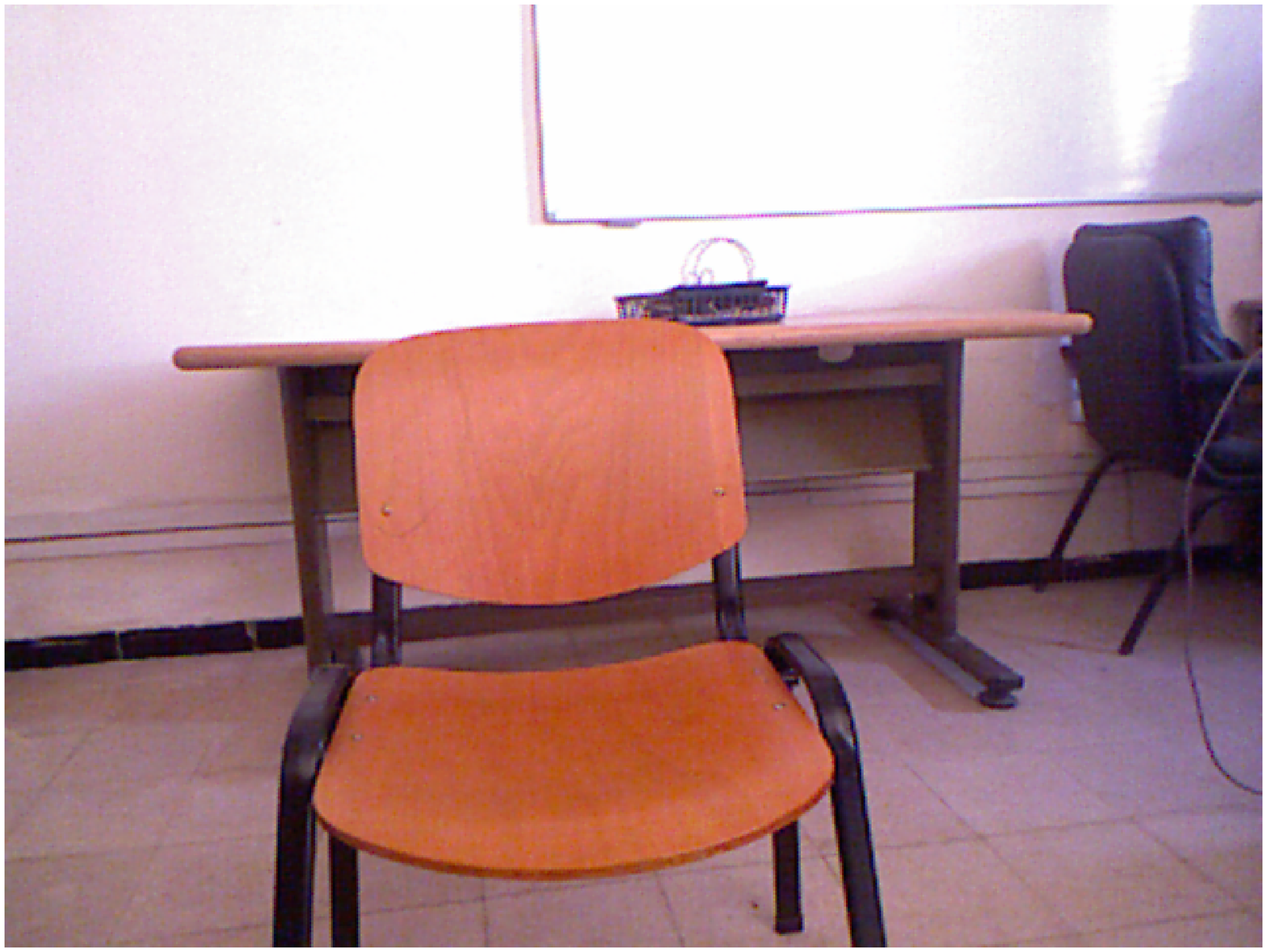}
\includegraphics[width=2cm]{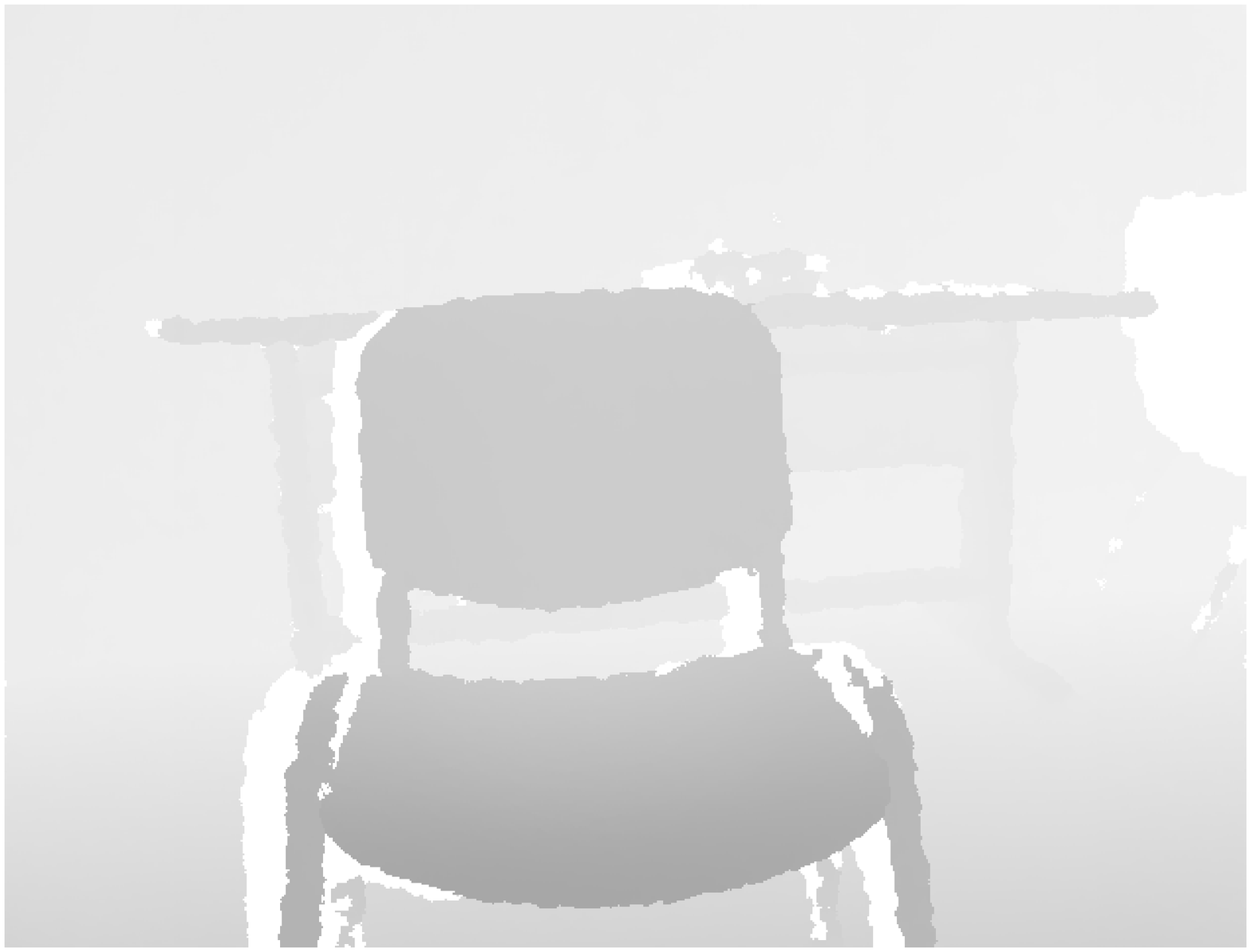}
\includegraphics[width=2cm]{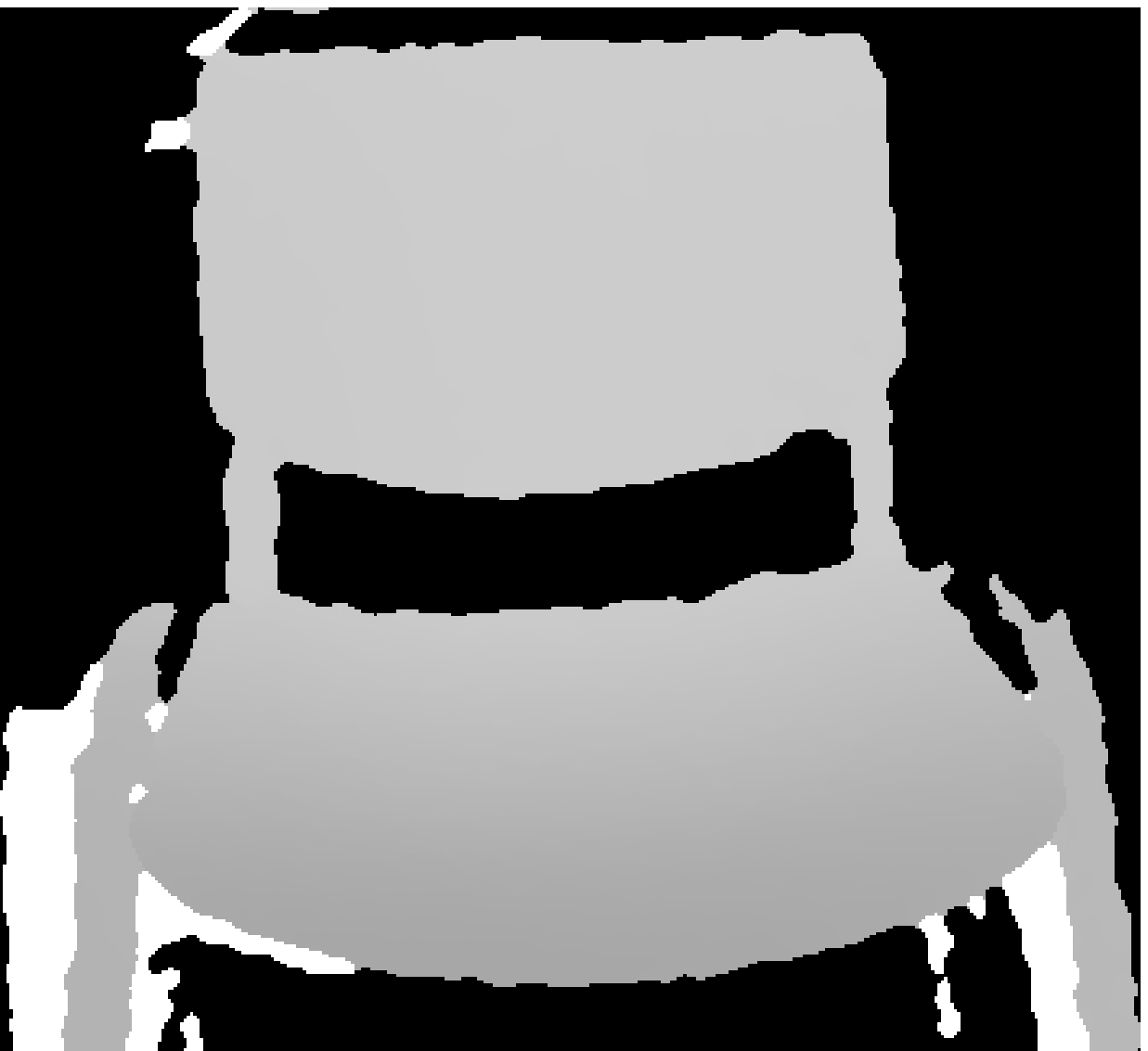}\\
\includegraphics[width=2cm]{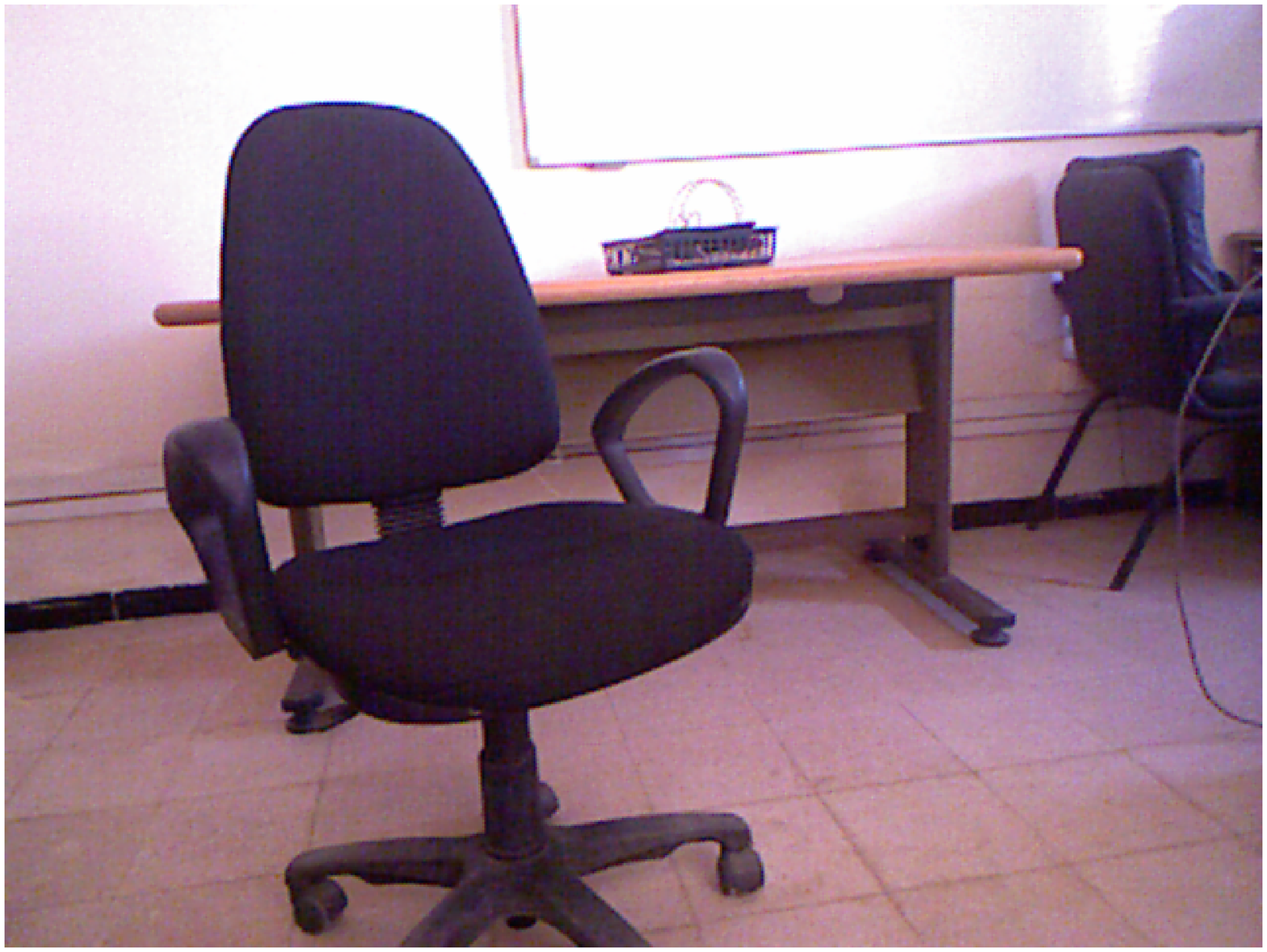}
\includegraphics[width=2cm]{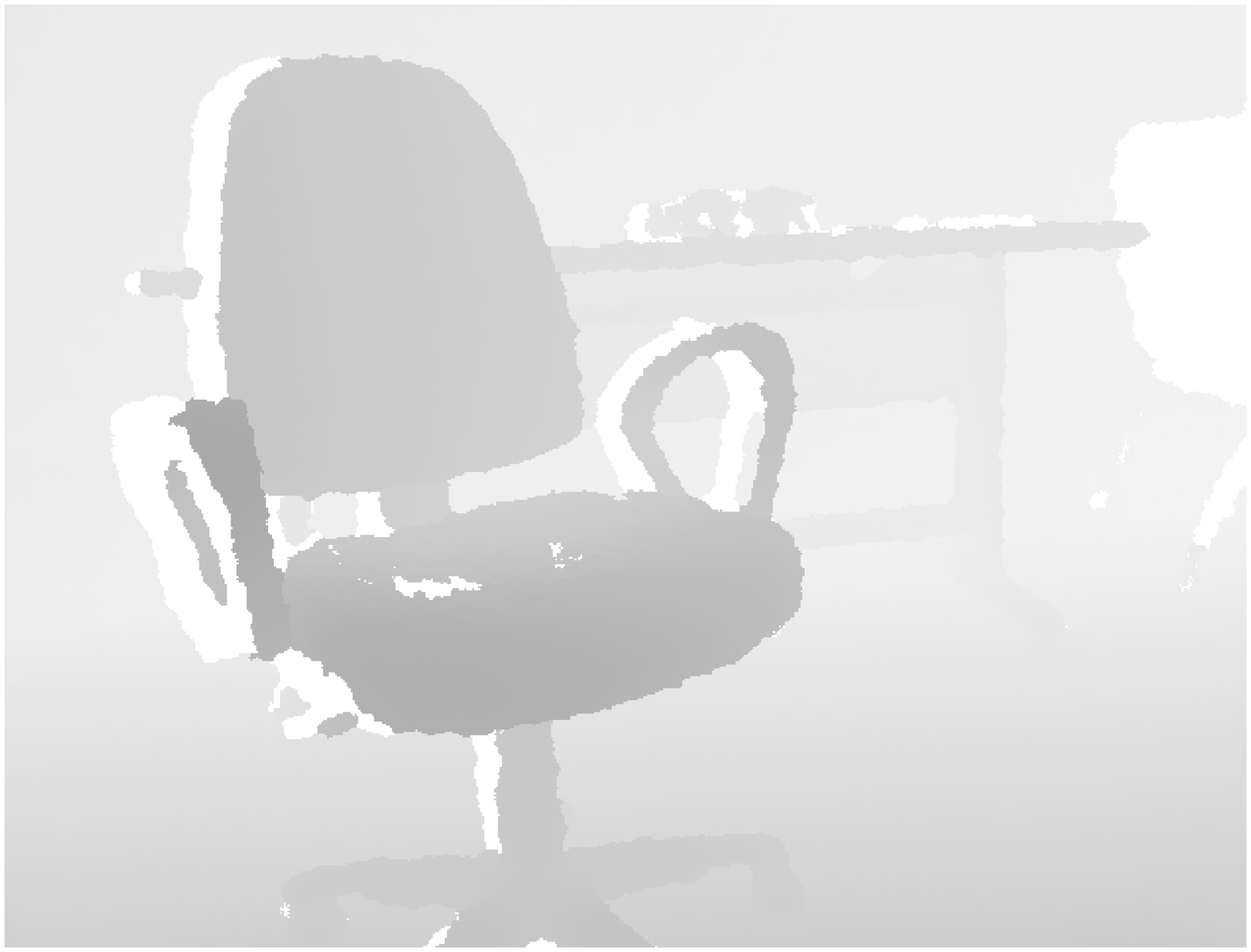}
\includegraphics[width=2cm]{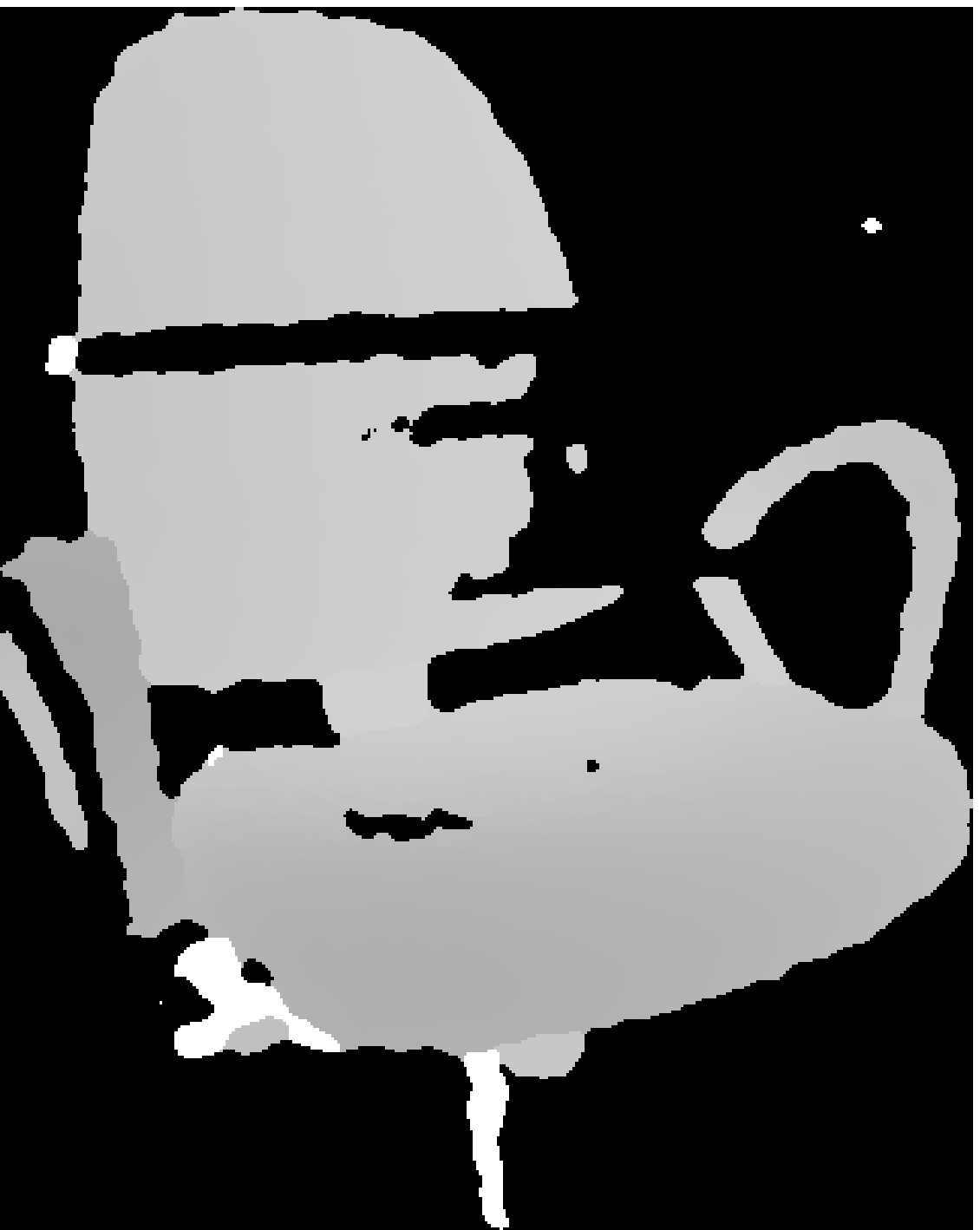}\\
\includegraphics[width=2cm]{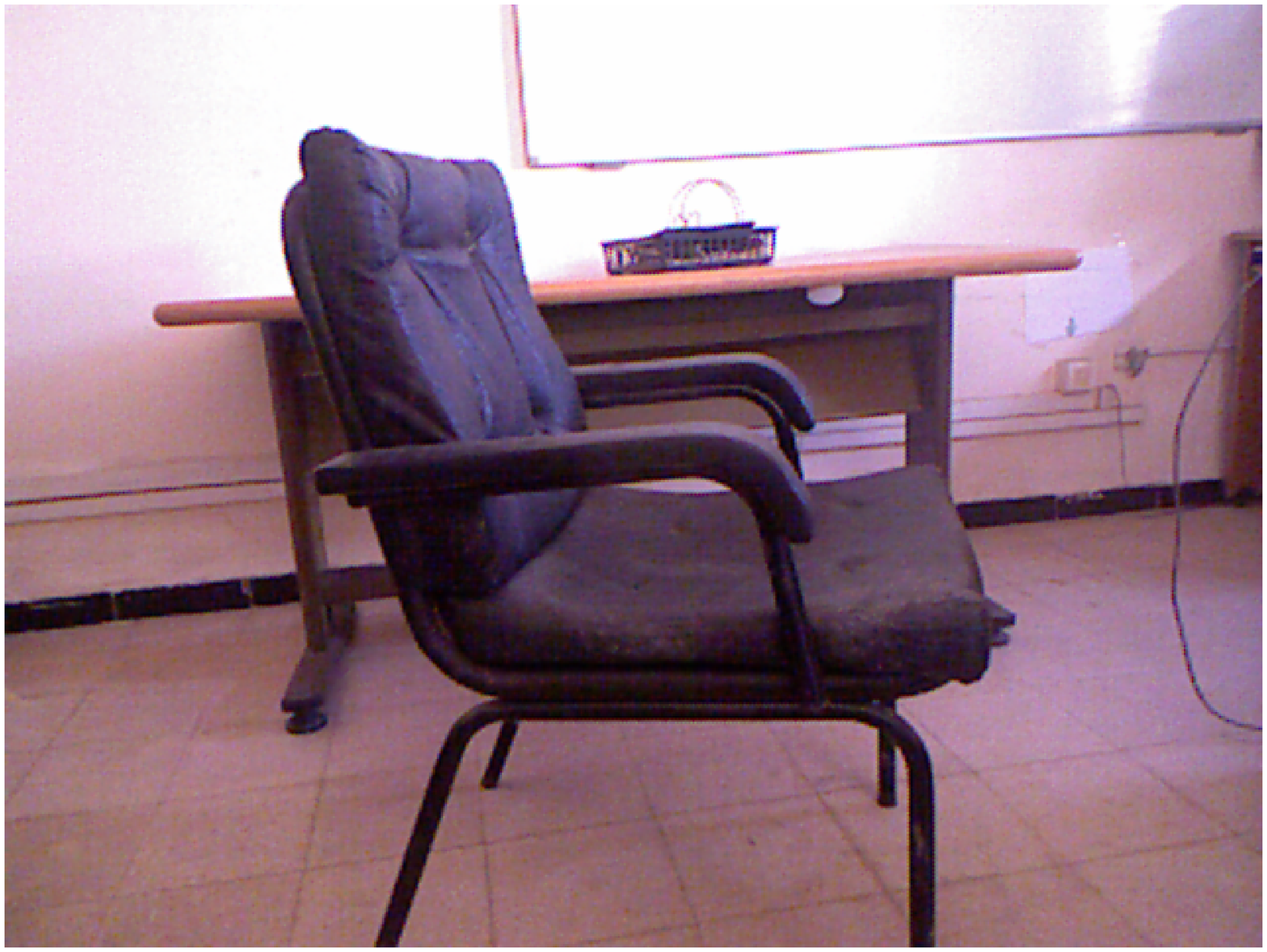}
\includegraphics[width=2cm]{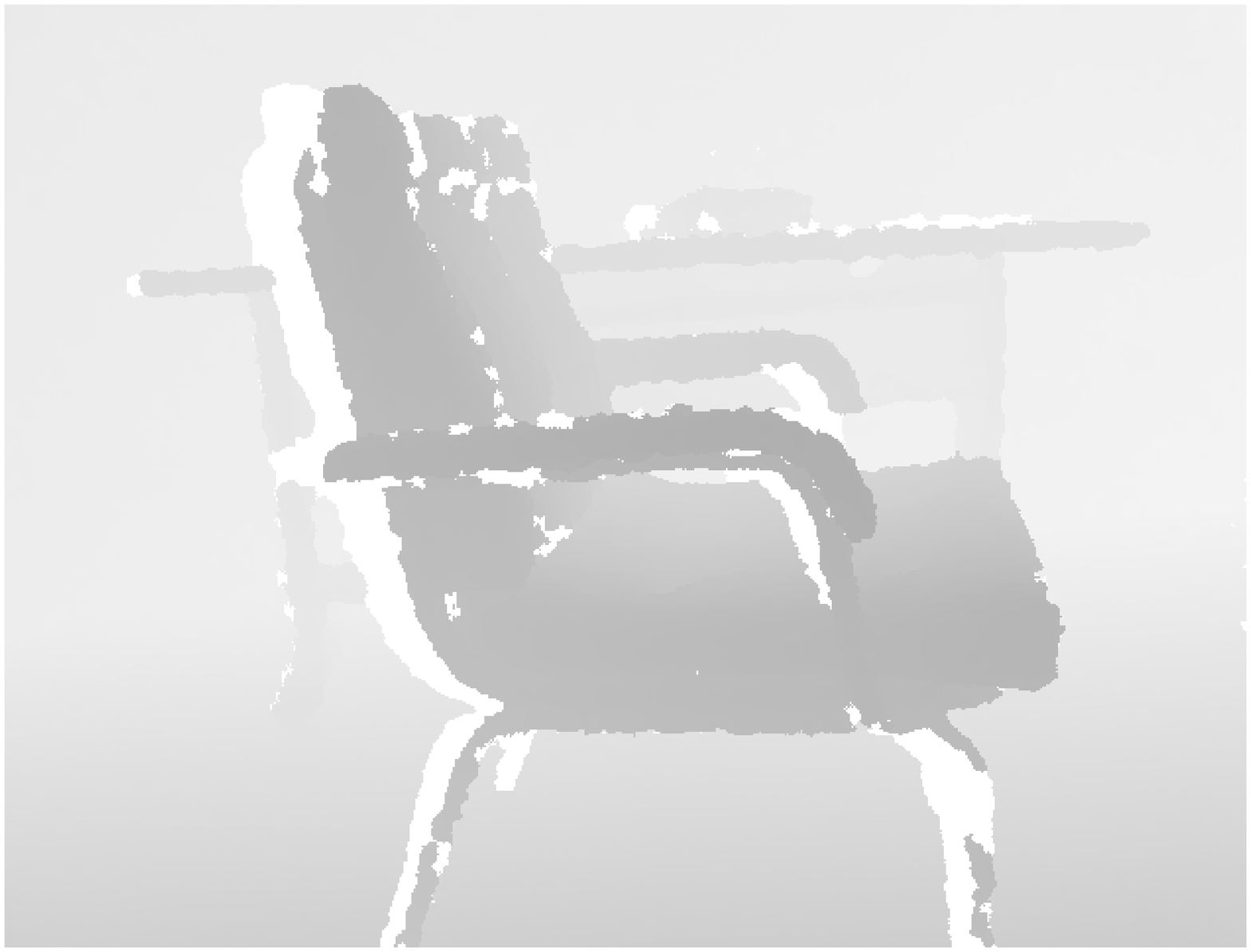}
\includegraphics[width=2cm]{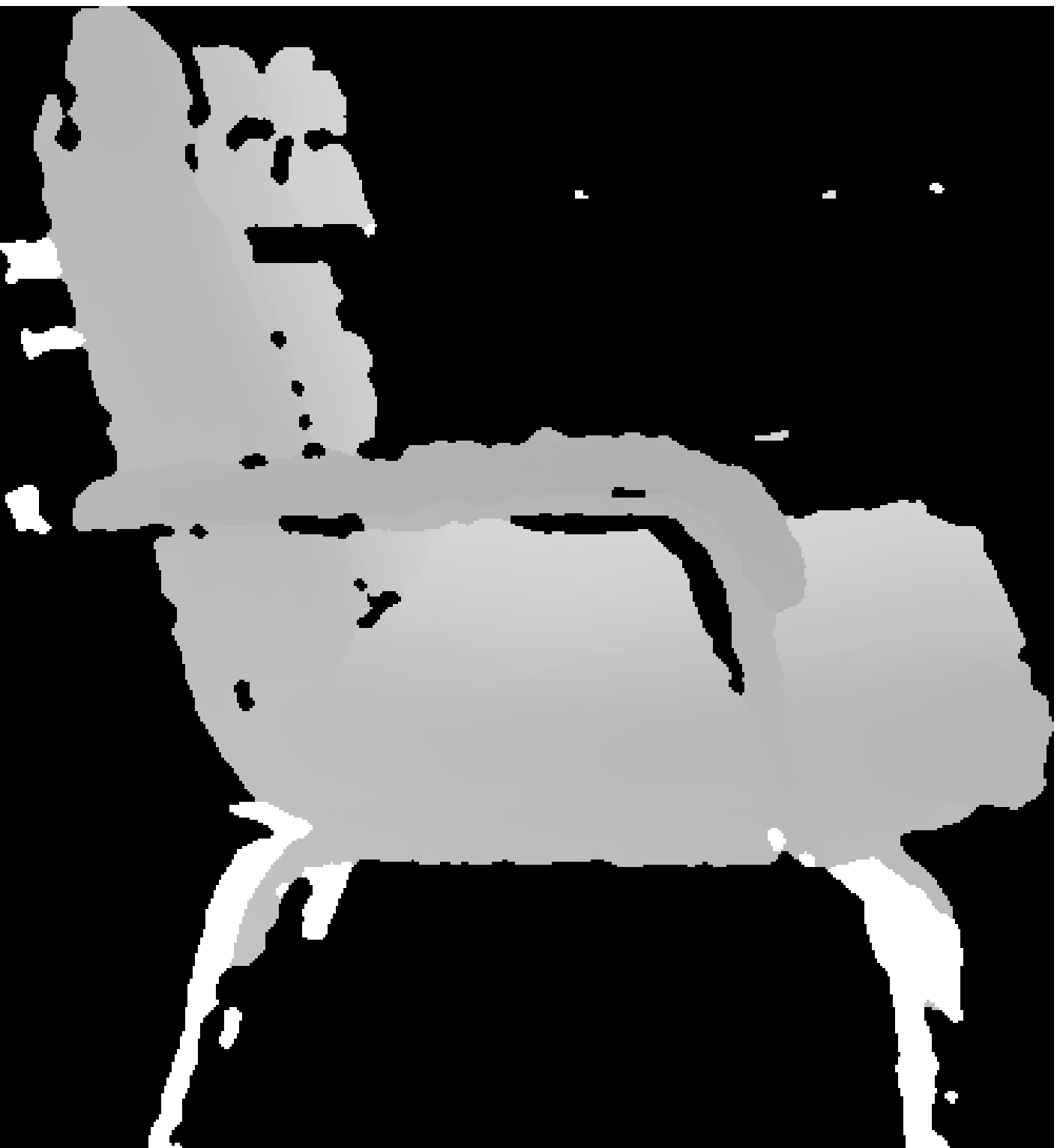}
	\caption{From left to right: the captured scene (chair, chair, chair), its associated depth image (the system input), the resulting segment. The system has predicted: chair, chair, chair.}
\label{fig:tests1}
\end{figure}

\begin{figure}[!t]
	\centering
\includegraphics[width=2.8cm]{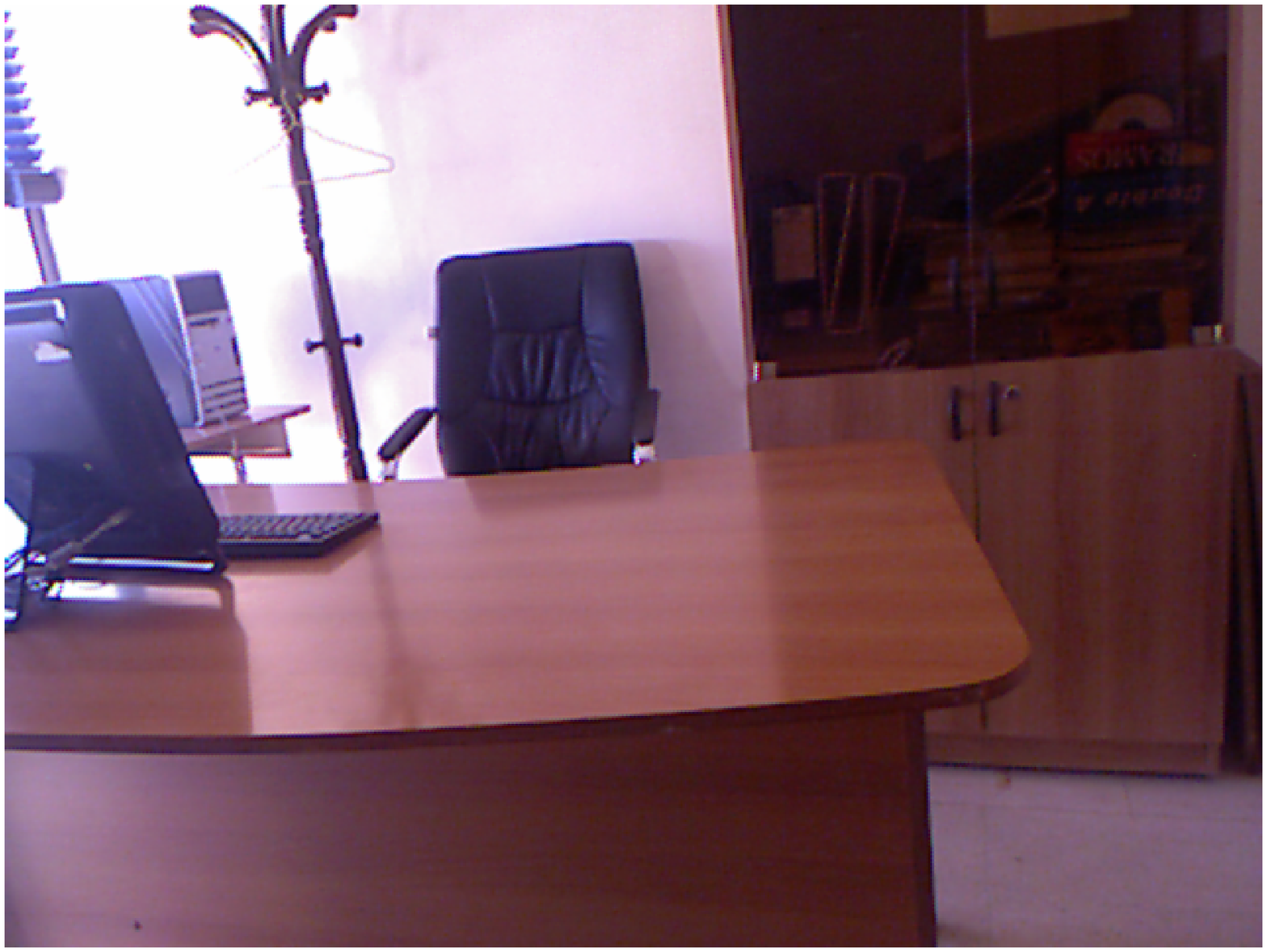}
\includegraphics[width=2.8cm]{rgb_table.eps}
\includegraphics[width=2.8cm]{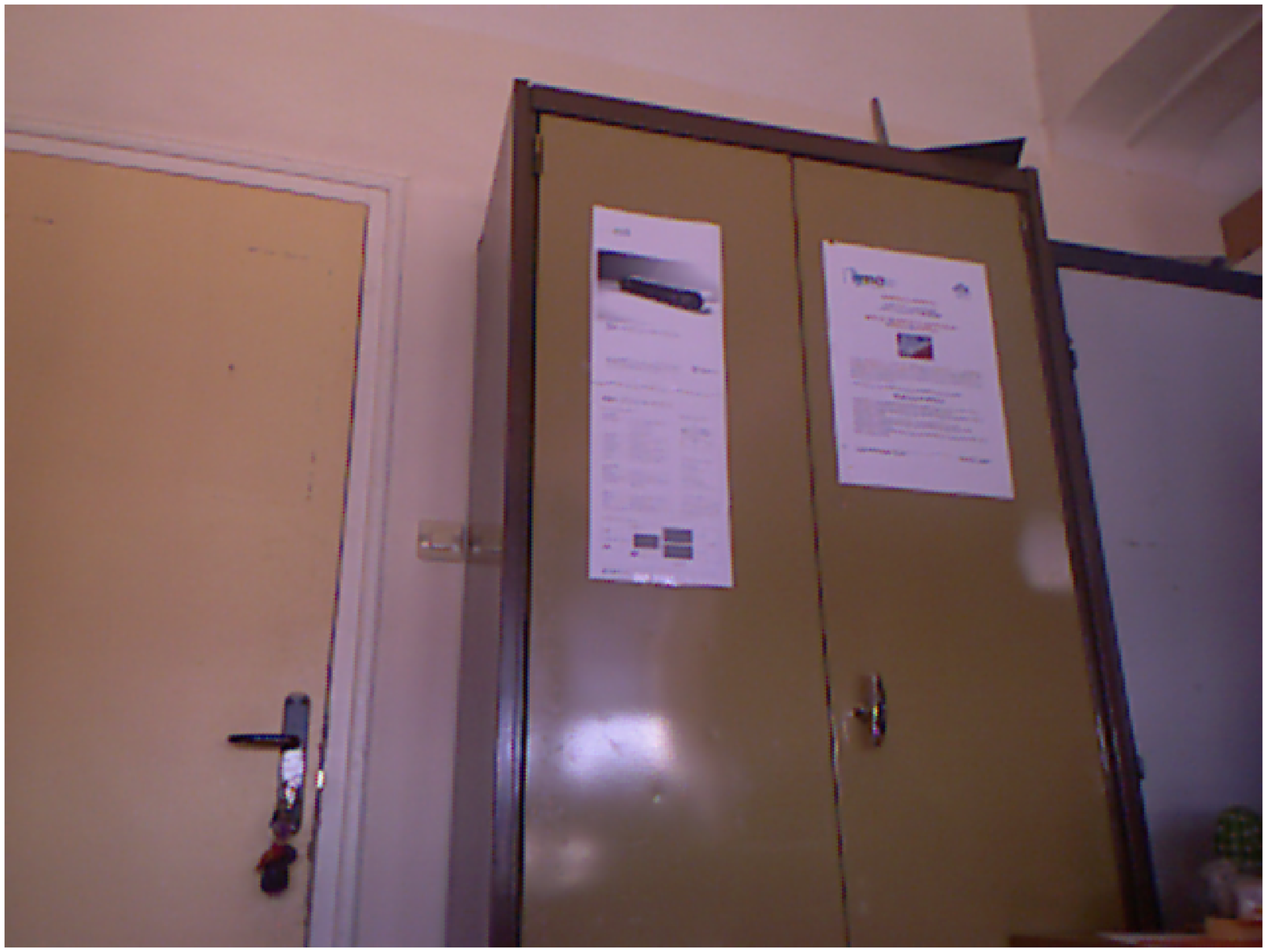}\\
\vspace{0.2cm}
\includegraphics[width=2.8cm]{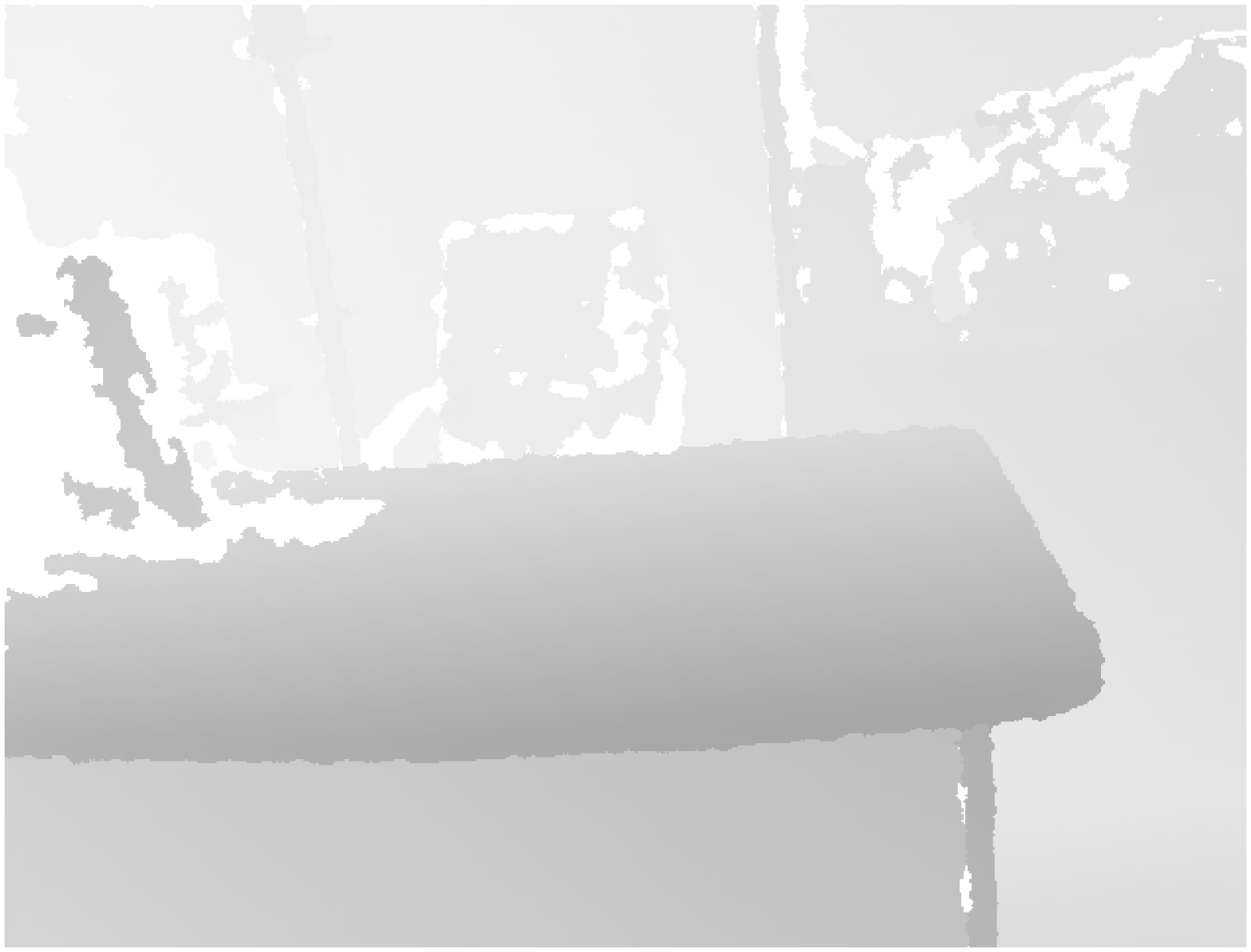}
\includegraphics[width=2.8cm]{p_table.eps}
\includegraphics[width=2.8cm]{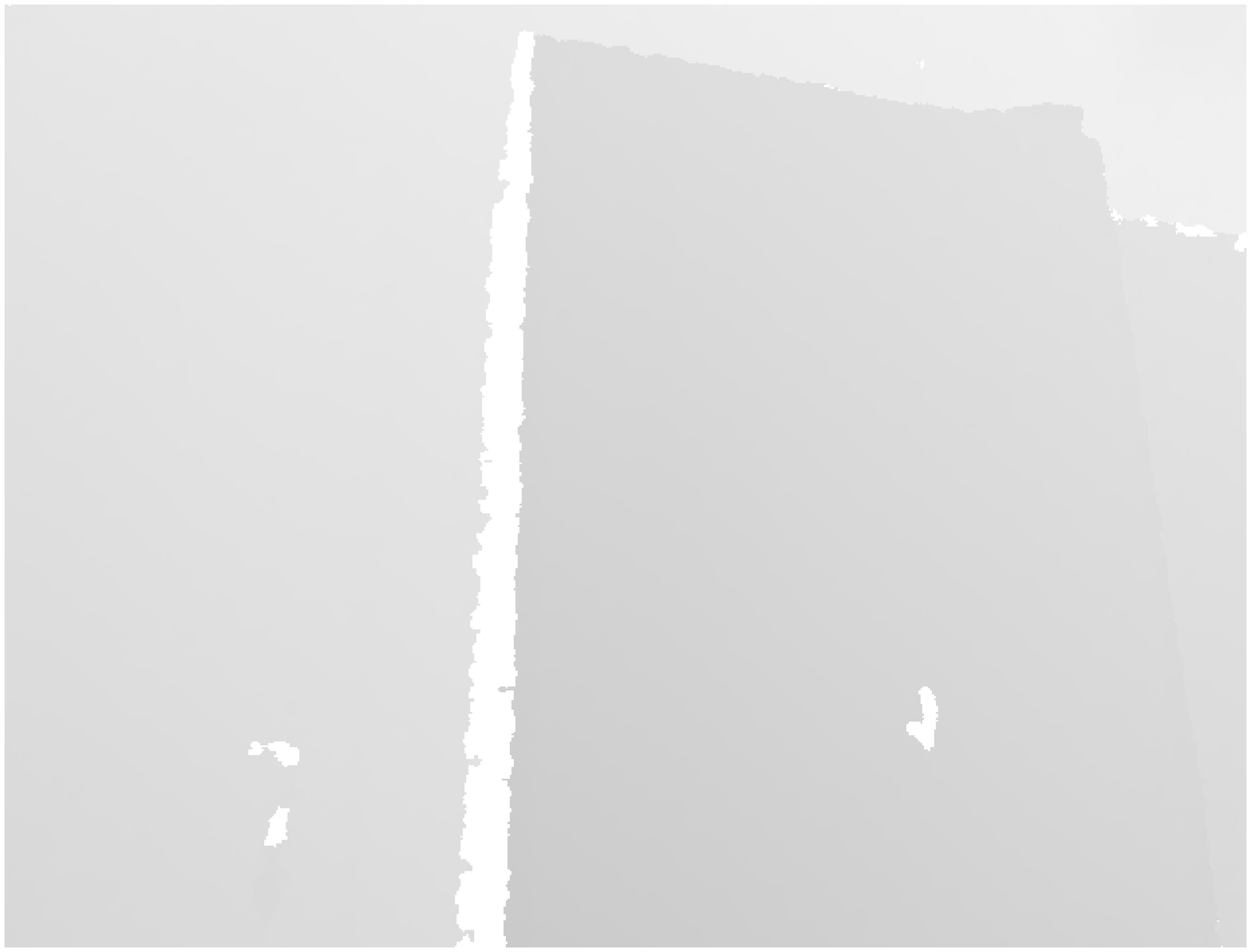}\\
\vspace{0.2cm}
\includegraphics[width=3.2cm]{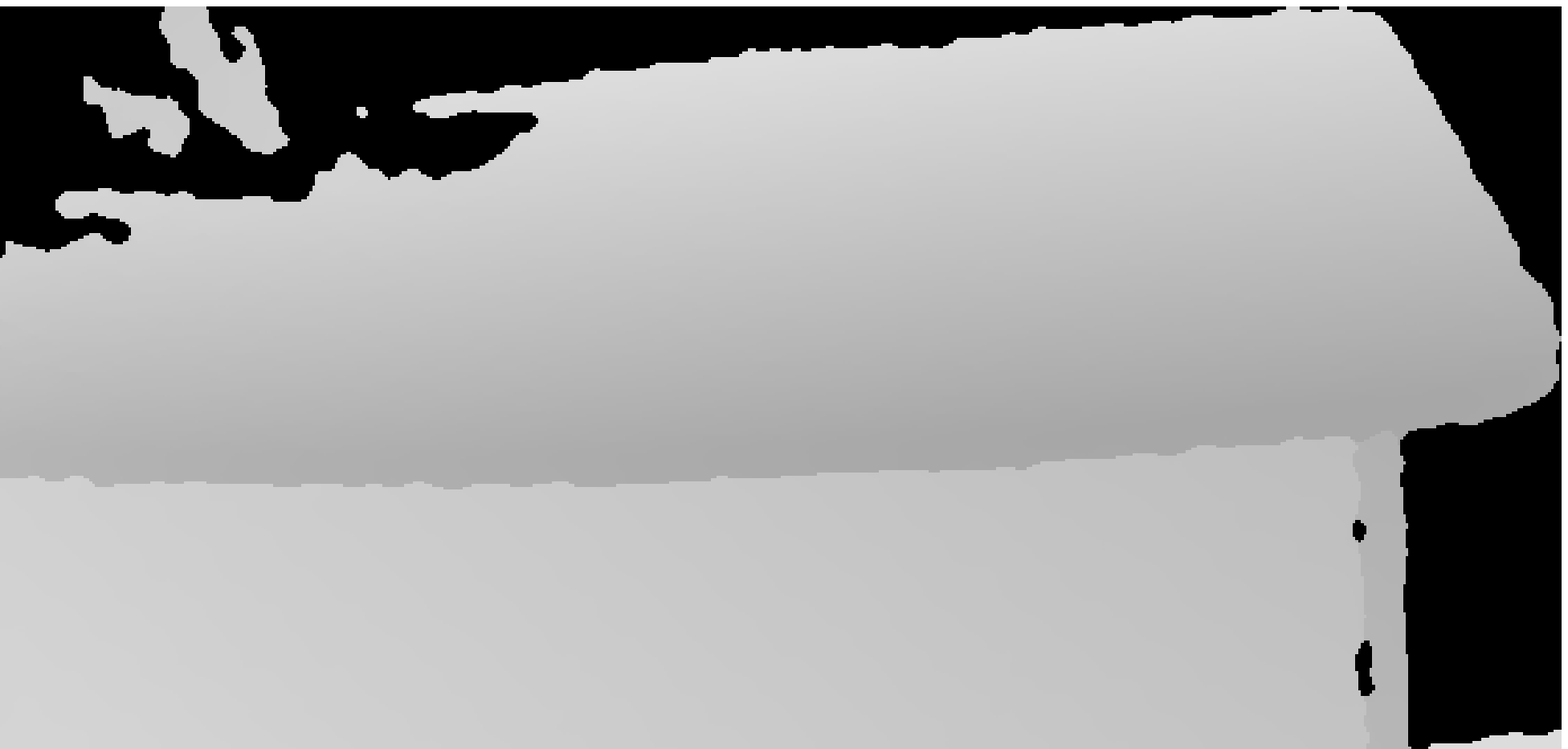}
\includegraphics[width=3.2cm]{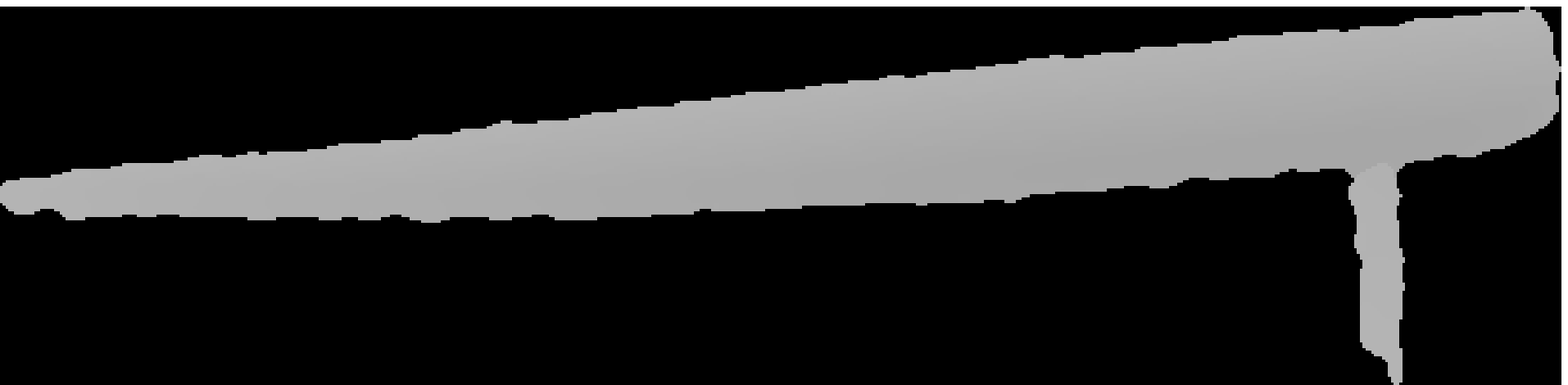}
\includegraphics[height=1.5cm]{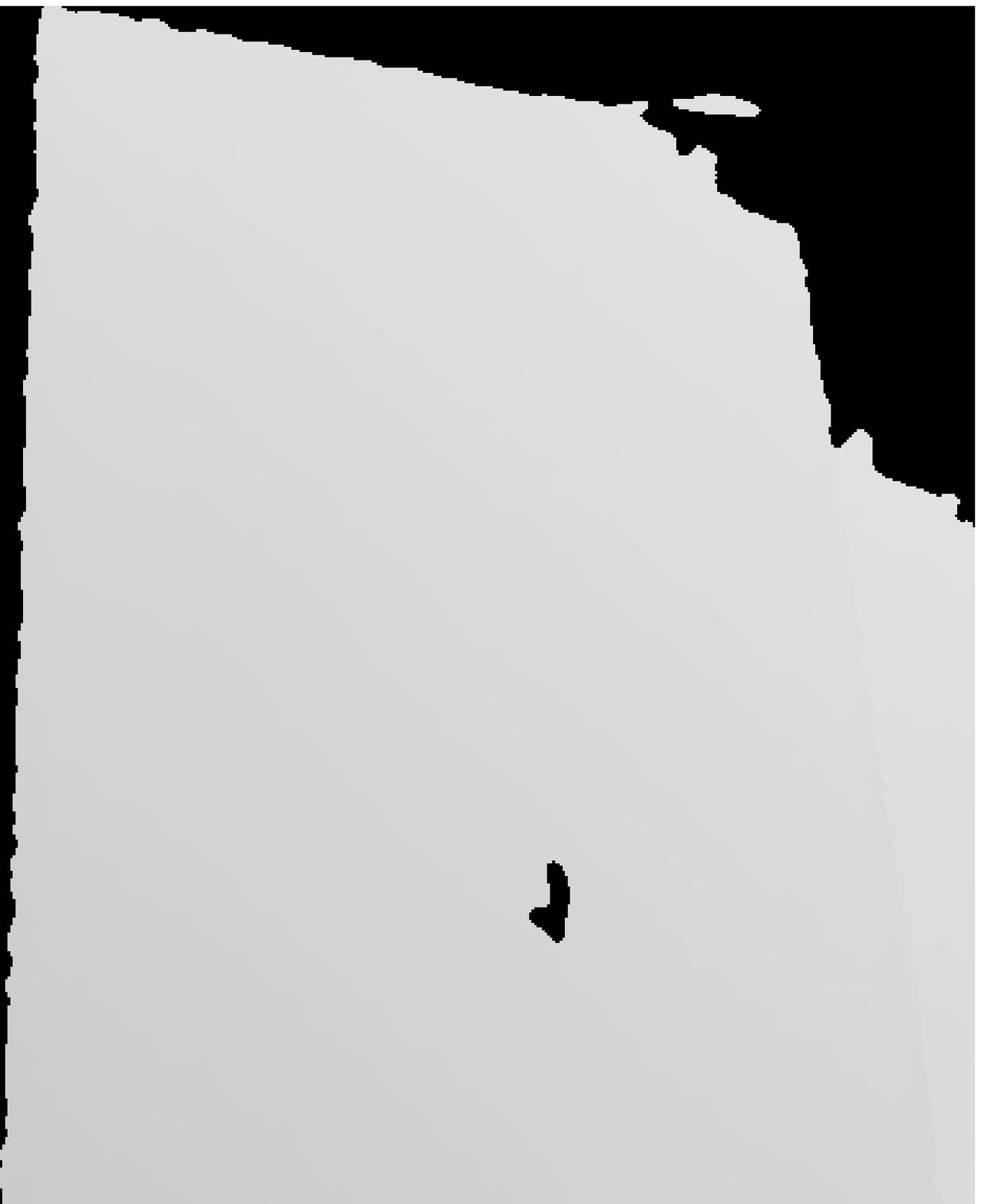}\\
	\caption{From top to down: the captured scene (table, table, dresser), its associated depth image (the system input), the resulting segment. The system has predicted: table, chair and dresser.}
\label{fig:tests2}
\end{figure}

The model was trained using 3D CAD models; on real-world data, the model predicts the right class when the input is noisy and slightly incomplete (see Figure \ref{fig:tests1}); however, it fails otherwise. As shown in Figure \ref{fig:tests2} (column 2), the system has predicted incorrectly the class of the table that was poorly segmented (we only used background removal to simulate the cropped data). To overcome this problem, we only consider the predicted class with high probability (higher than 0.85).

Compared with the results of object classification based aid systems presented previously (Table \ref{tab:comparison}), in the system \cite{wang2017enabling}, authors didn't mention the accuracy of their proposed depth-based classification algorithm. As for the system \cite{lin2019deep}, authors adapted FuseNet for RGB-D semantic segmentation that has 76.27\% as accuracy. Compared to the third system \cite{wang2014segment}, our model surpasses its accuracy and with a larger set of classes.
\begin{table}[ht]
	\centering
	\caption{Comparison with the classification module of the presented state of the art systems.}
	\begin{tabular}[t]{|c|c|c|c|c|}
	\hline
	 & Input &  Nb Classes & Classifier & Accuracy (\%)\\
	\hline
	Wang et al.\cite{wang2017enabling}&Depth image&5& Depth-based classifier\cite{wang2017enabling}& -\\
	\hline
	Lin et al.\cite{lin2019deep}      &  RGB-D      &  70& FuseNet\cite{hazirbas2016fusenet}& -\\
	\hline
	Wang et al.\cite{wang2014segment} &  RGB-D      &   4& Cascaded Decision Tree& 71.45\\
	\hline
	Ours                   &  Point cloud&  14& PointNet\cite{qi2017pointnet}& 91.67\\
	Ours                   &  Point cloud&   6& PointNet\cite{qi2017pointnet}& 96.35\\
	\hline
	\end{tabular}
	\label{tab:comparison}
\end{table}
%%%%%%%%%%%%%%%%%%%%%%%%%%%%%%%%%%%%%%%%  TESTS %%%%%%%%%%%%%%%%%%%%%%%%%%%%%%%%%%%%%%%%%%%%%%%%%%%%%%
\subsection{Tests on GDIS dataset \cite{GDIS2019}}
\begin{figure}[t]
	\centering
		\includegraphics[width=2.7cm]{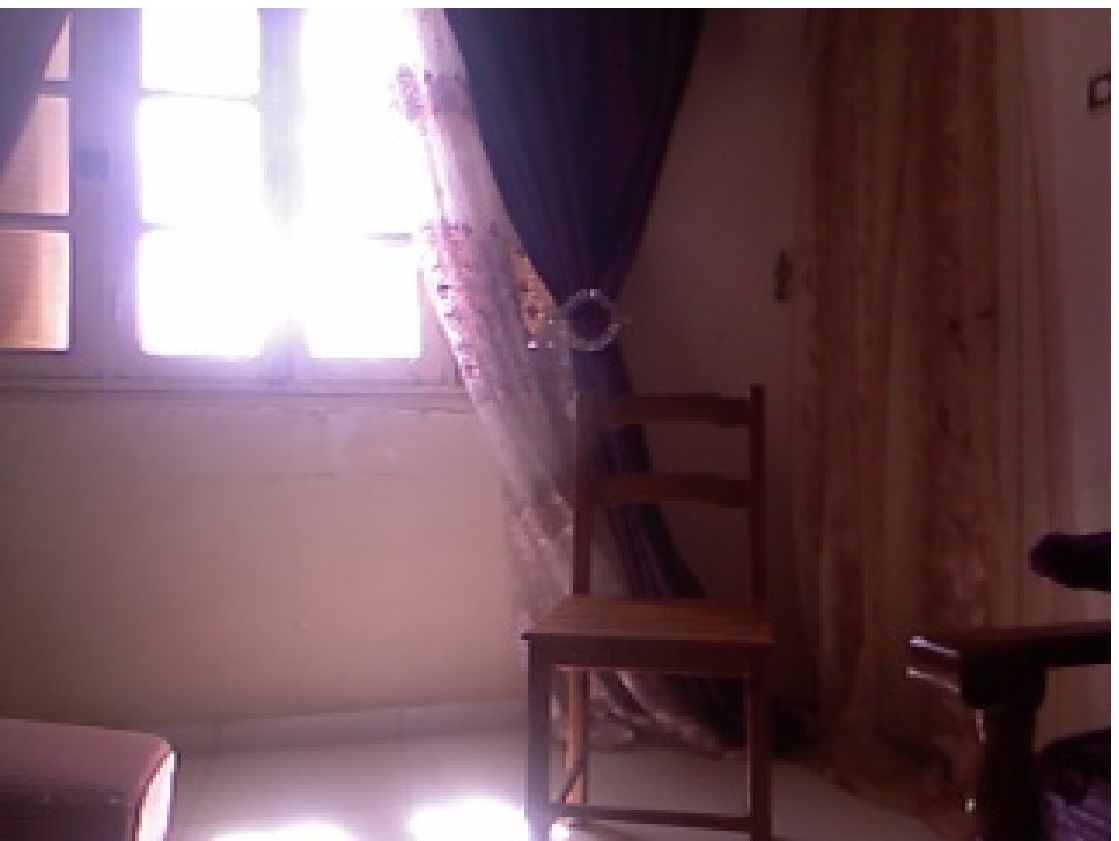}
		\includegraphics[width=2.7cm]{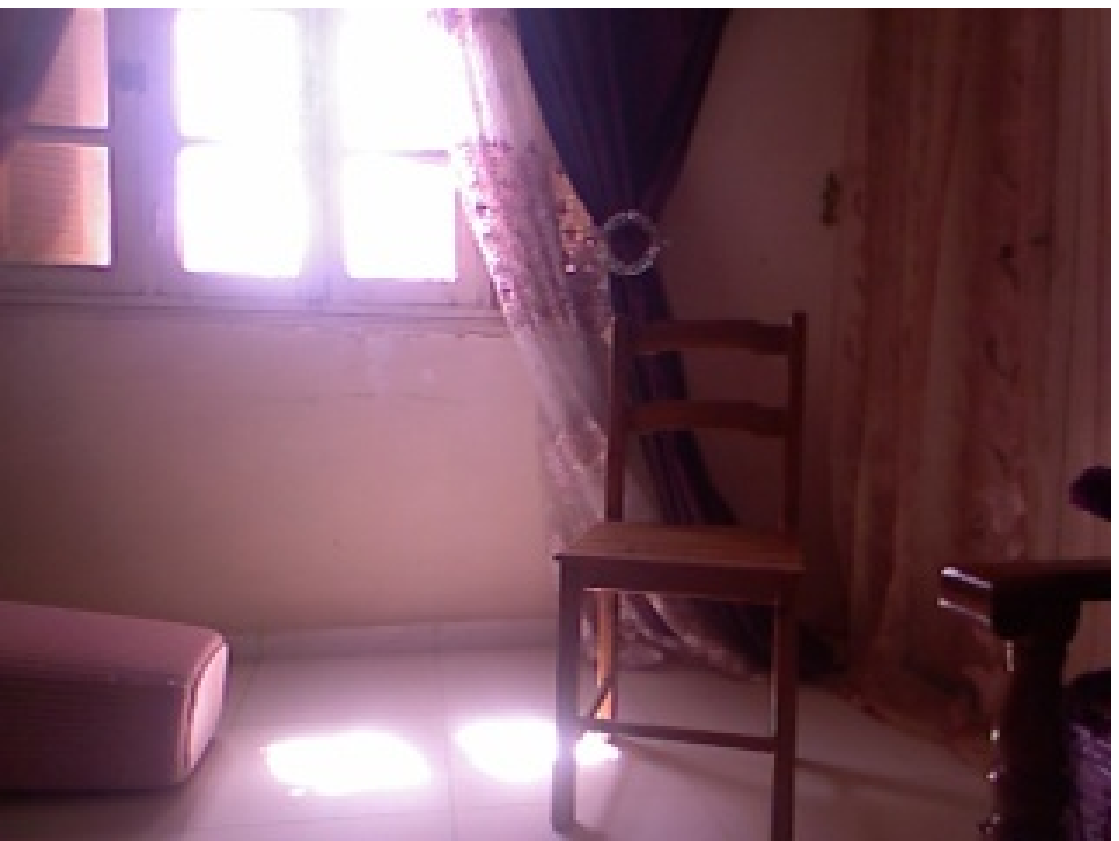}
		\includegraphics[width=2.7cm]{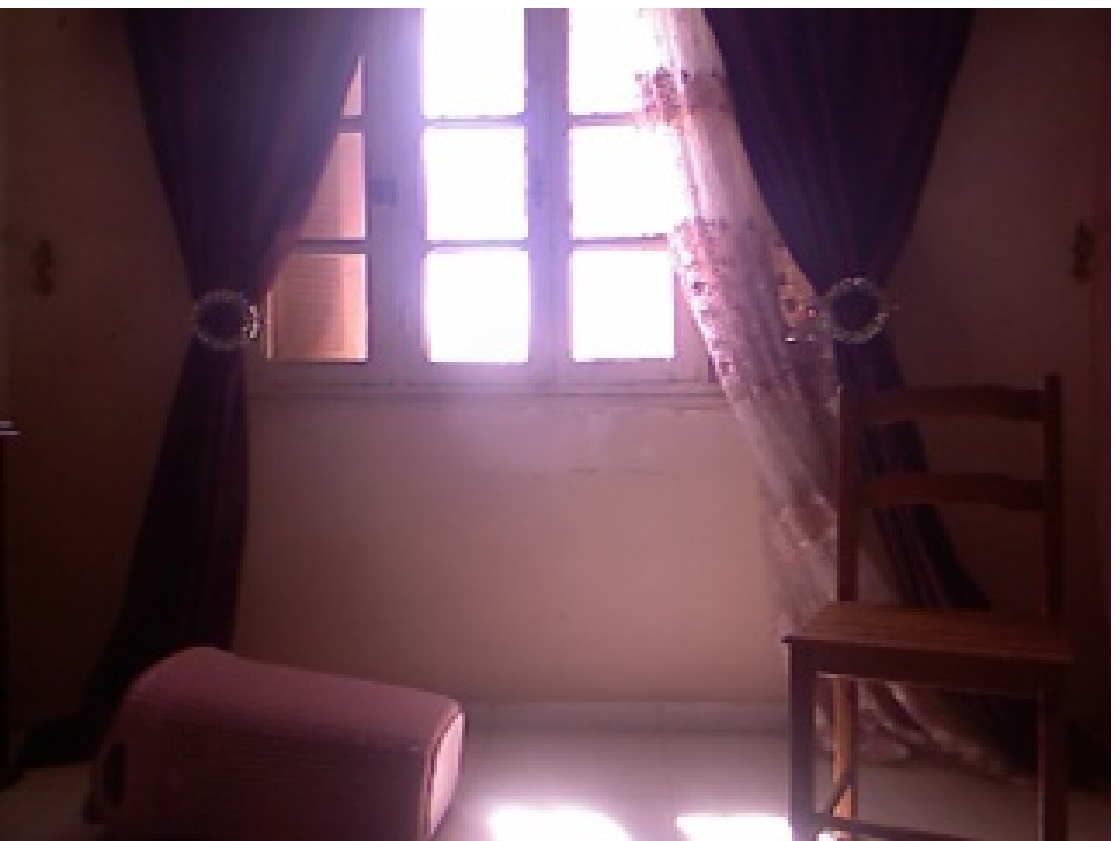}
		\includegraphics[width=2.7cm]{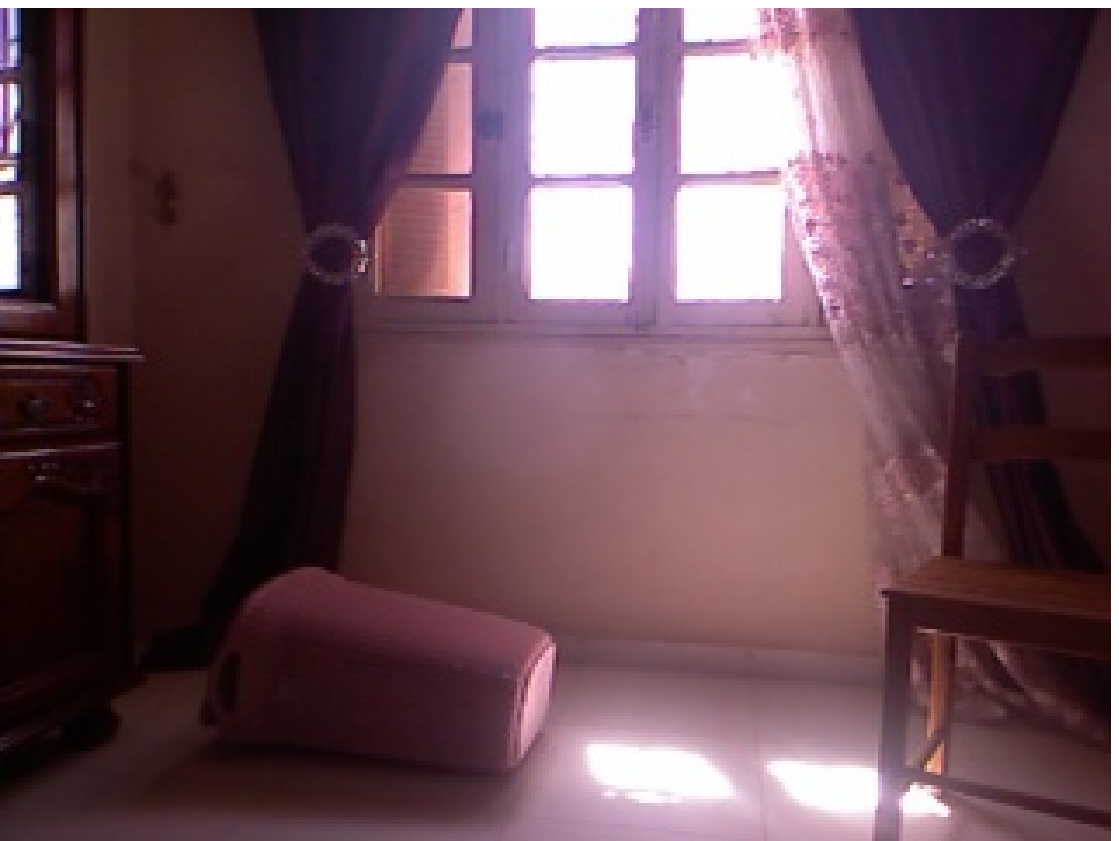}
		\includegraphics[width=2.7cm]{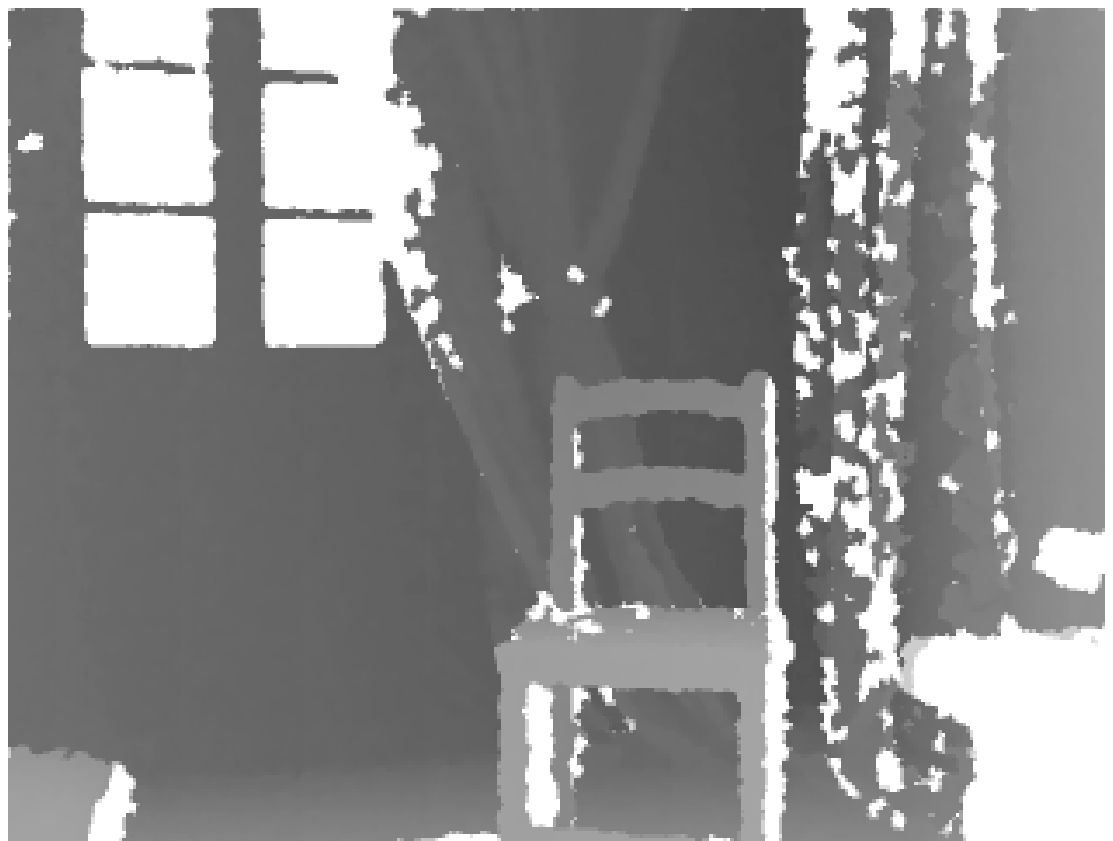}
		\includegraphics[width=2.7cm]{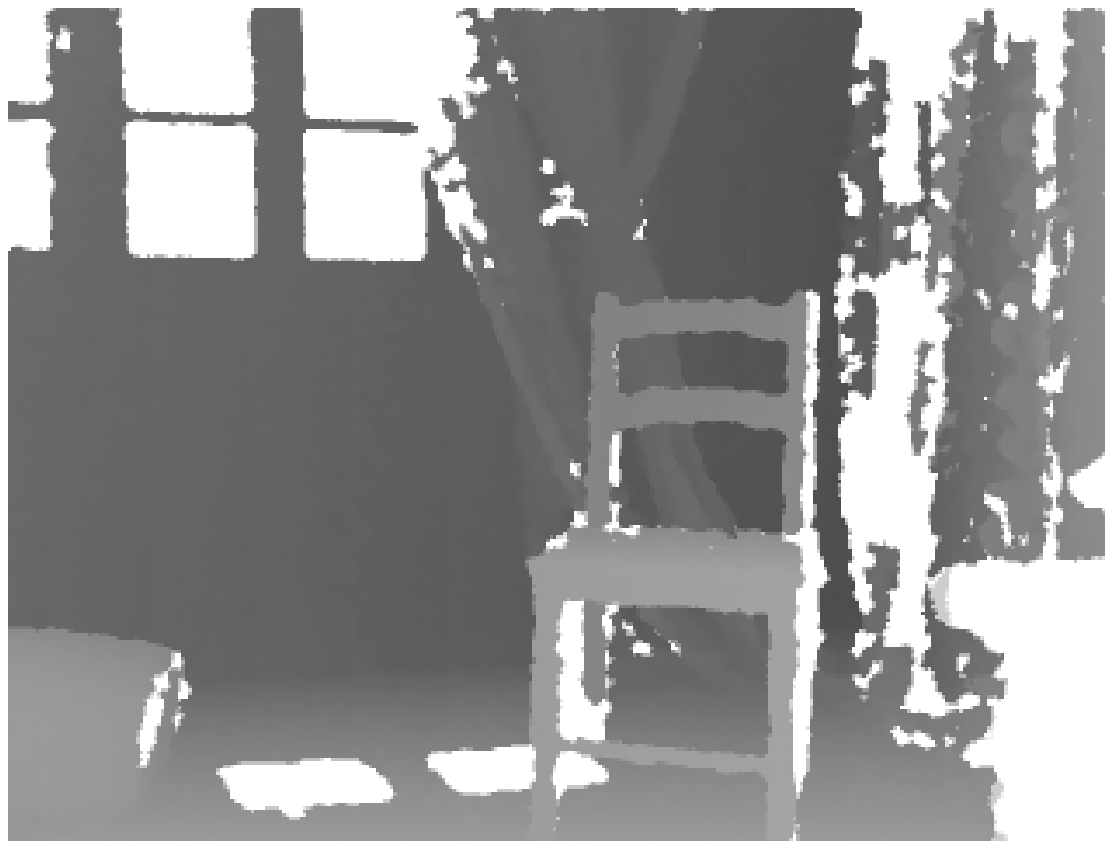}
		\includegraphics[width=2.7cm]{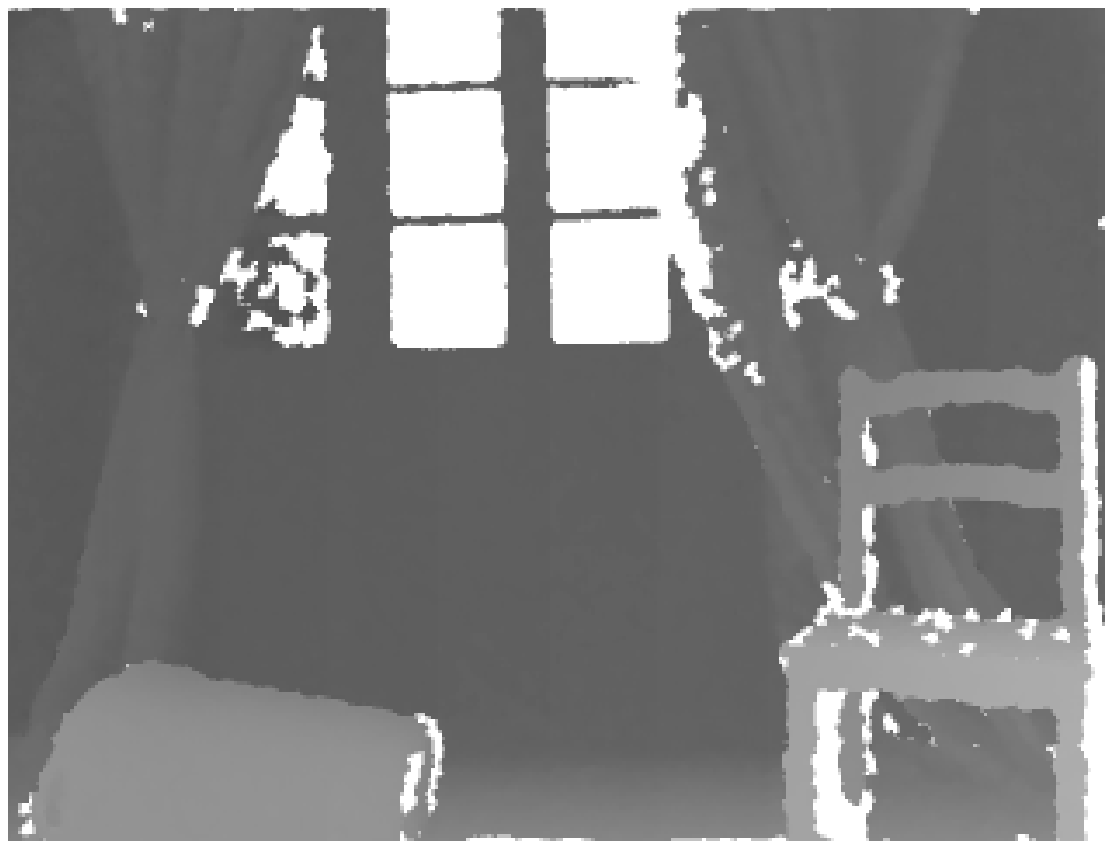}
		\includegraphics[width=2.7cm]{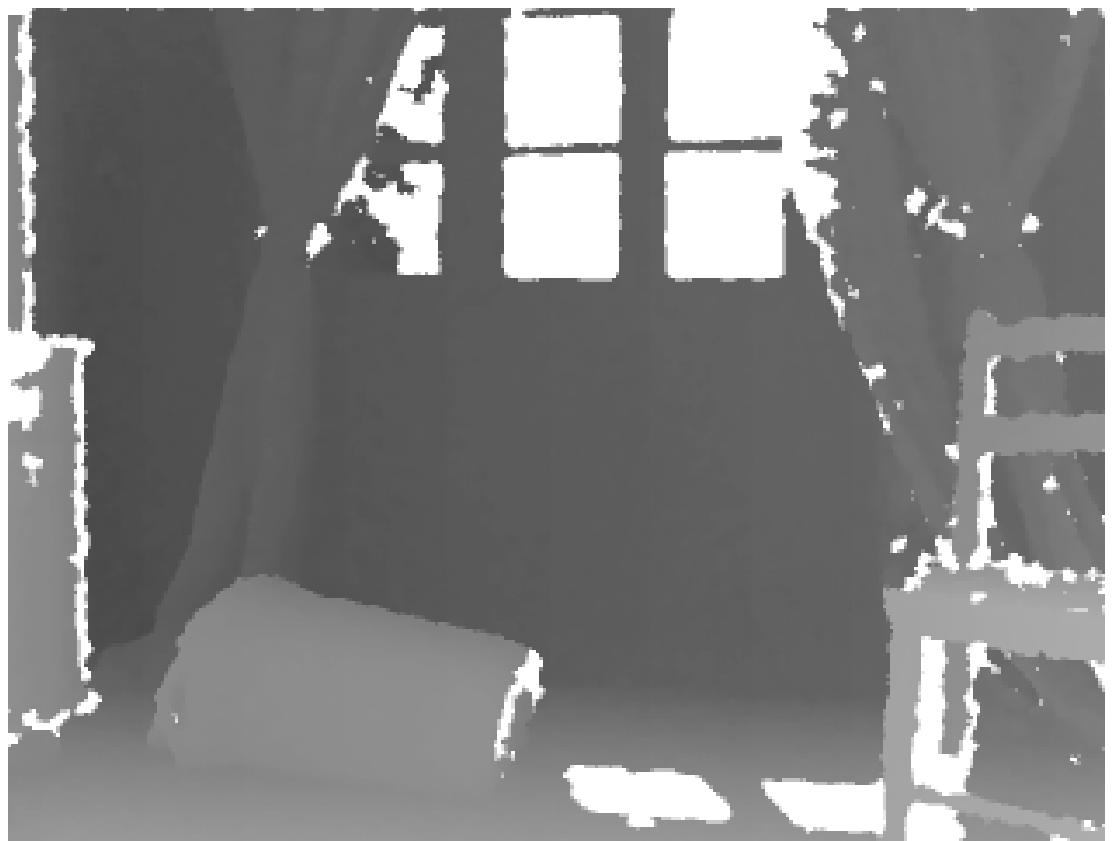}
		\includegraphics[width=2.7cm]{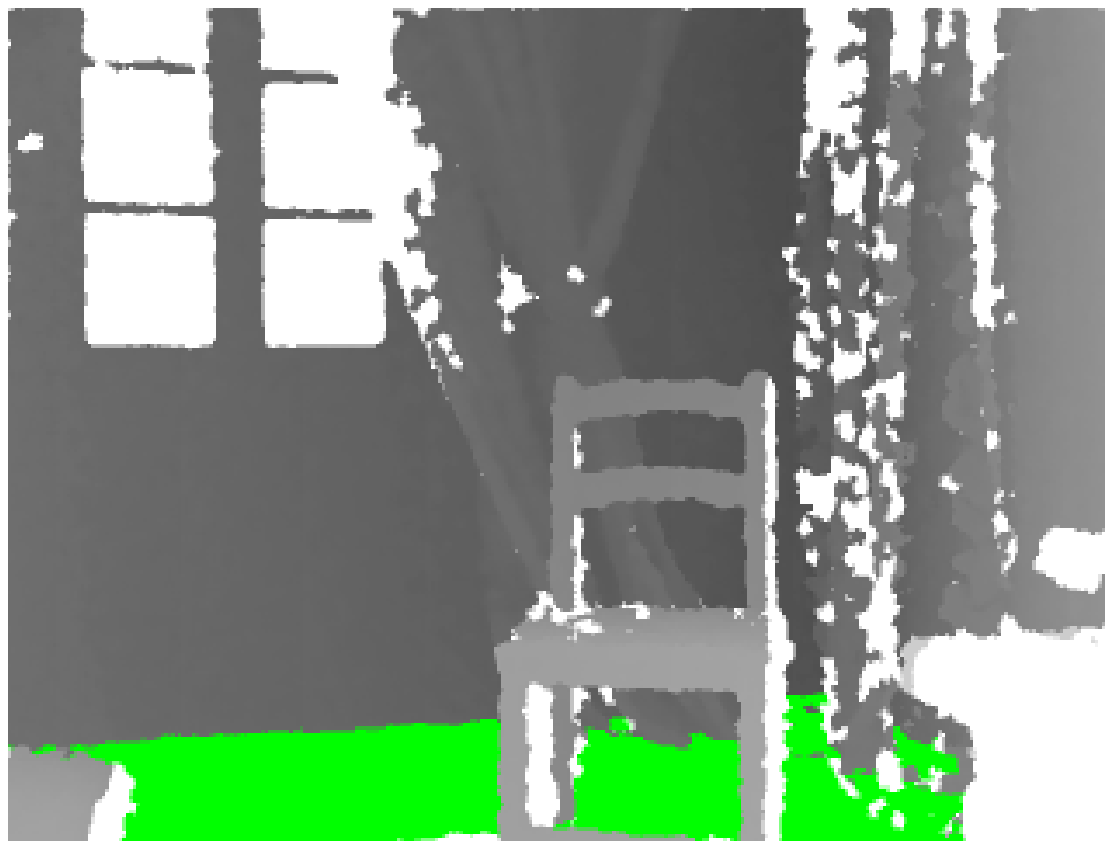}
		\includegraphics[width=2.7cm]{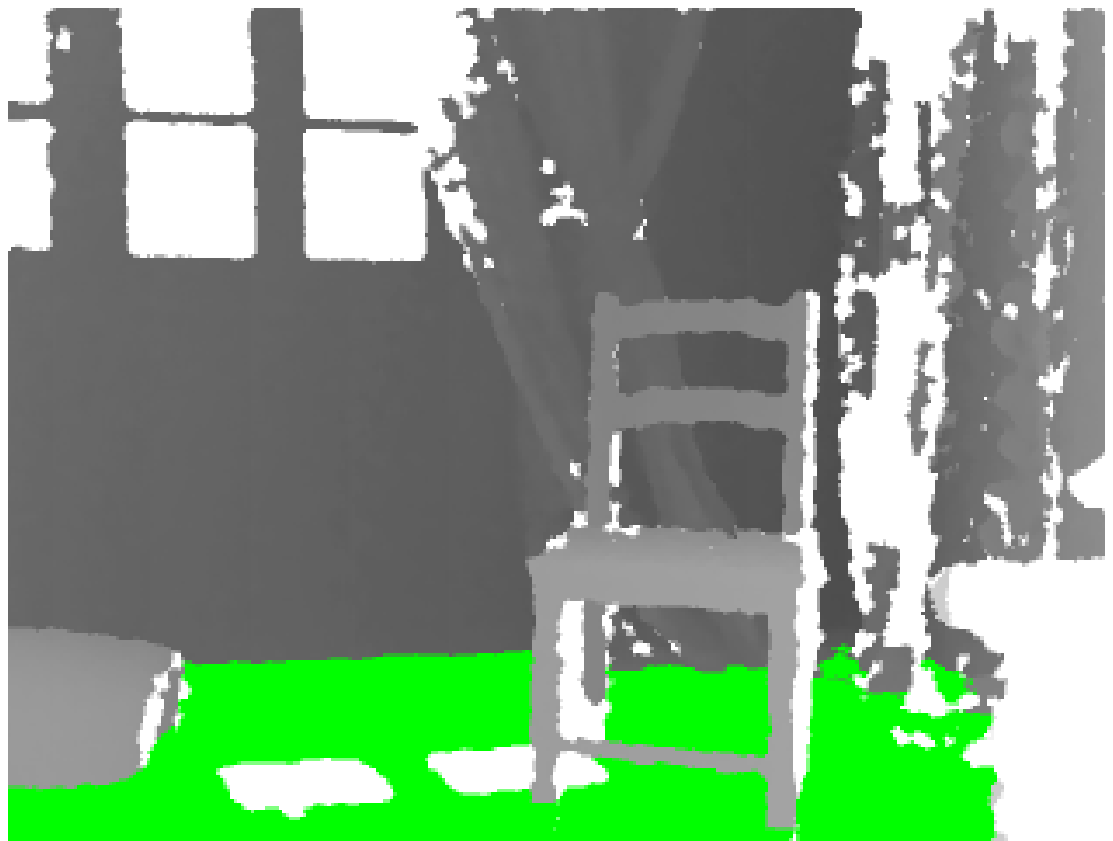}
		\includegraphics[width=2.7cm]{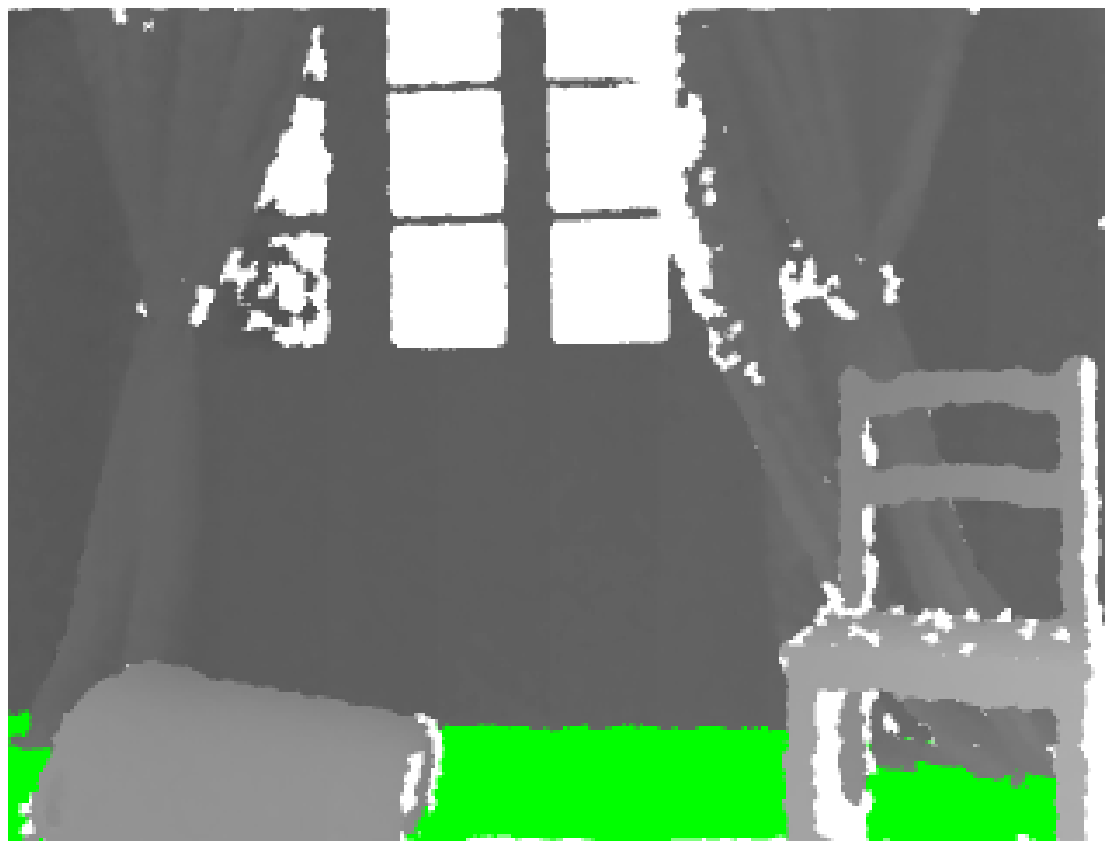}
		\includegraphics[width=2.7cm]{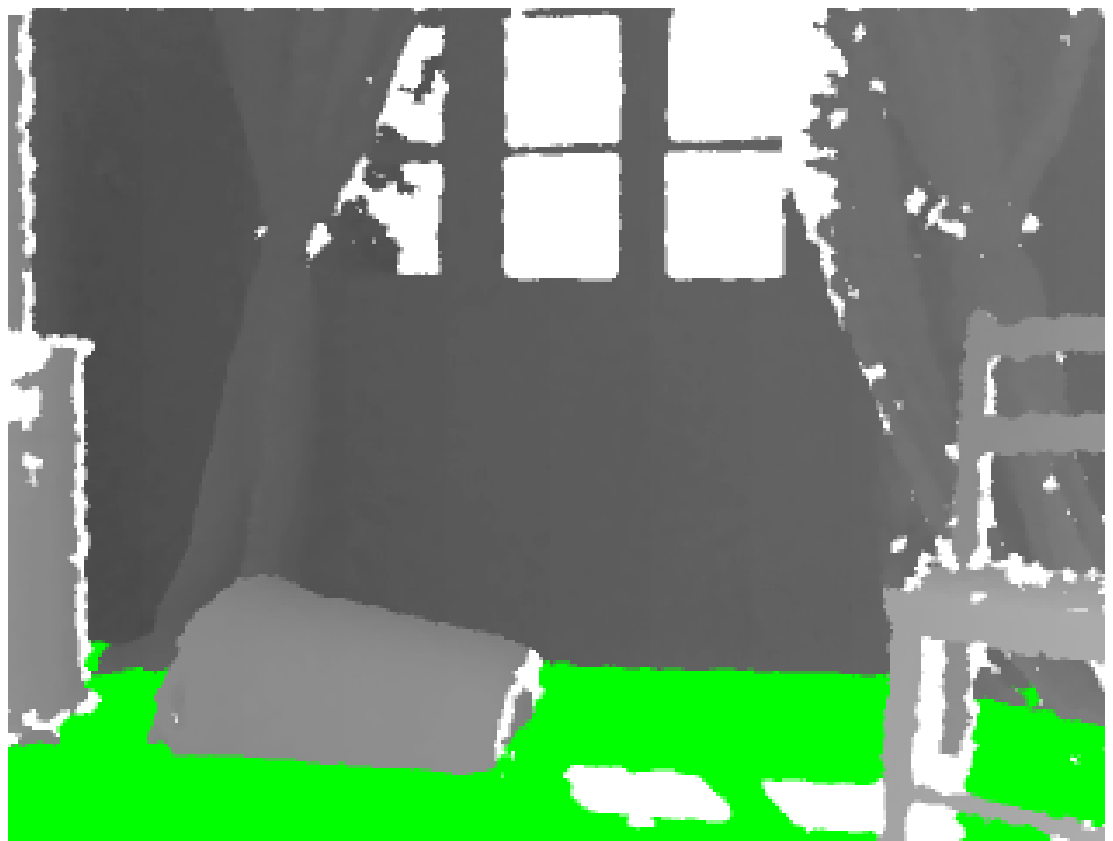}
		\includegraphics[width=2.7cm]{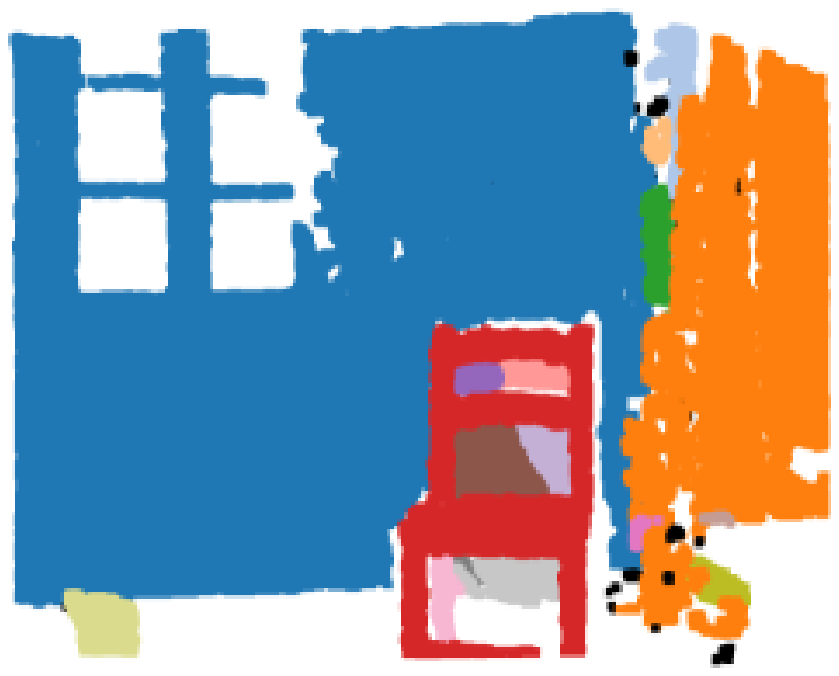}
		\includegraphics[width=2.7cm]{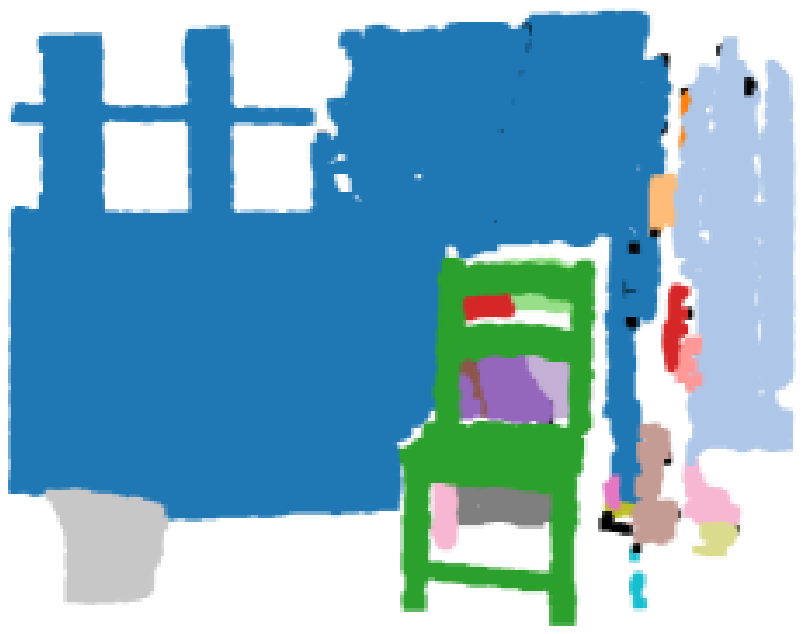}
		\includegraphics[width=2.7cm]{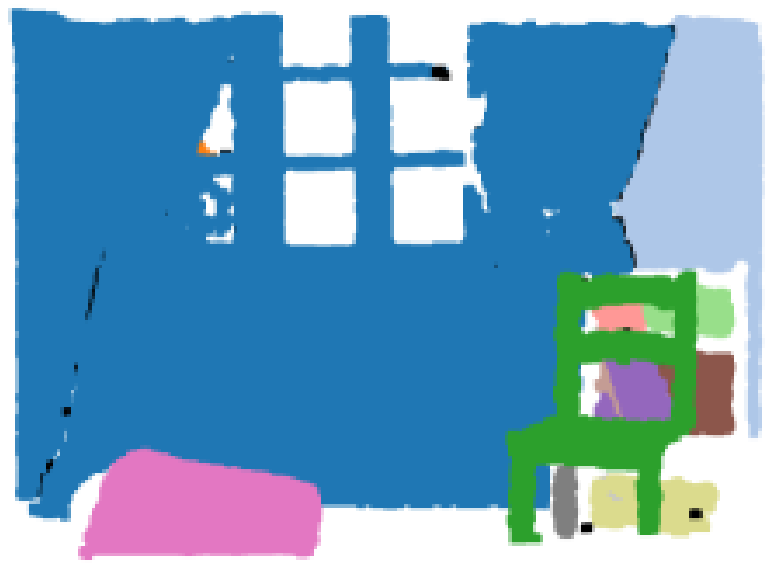}
		\includegraphics[width=2.7cm]{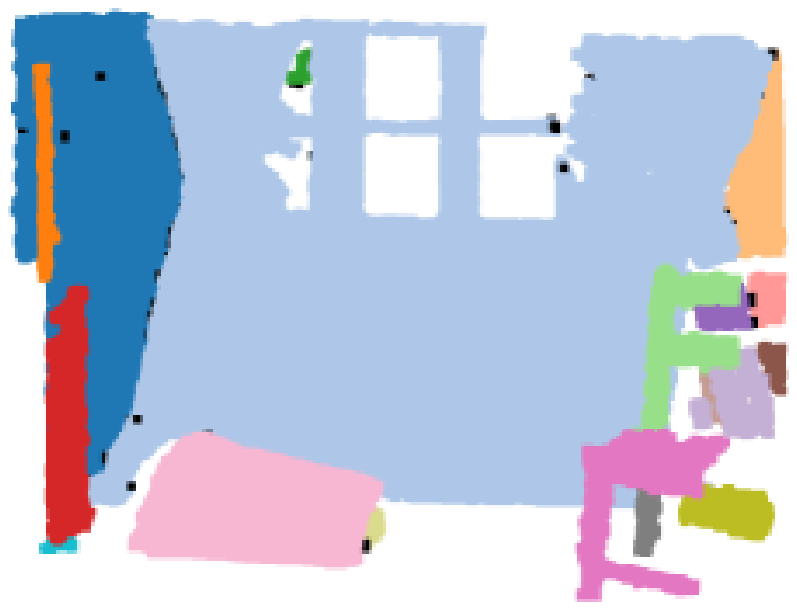}
		\includegraphics[width=2.7cm]{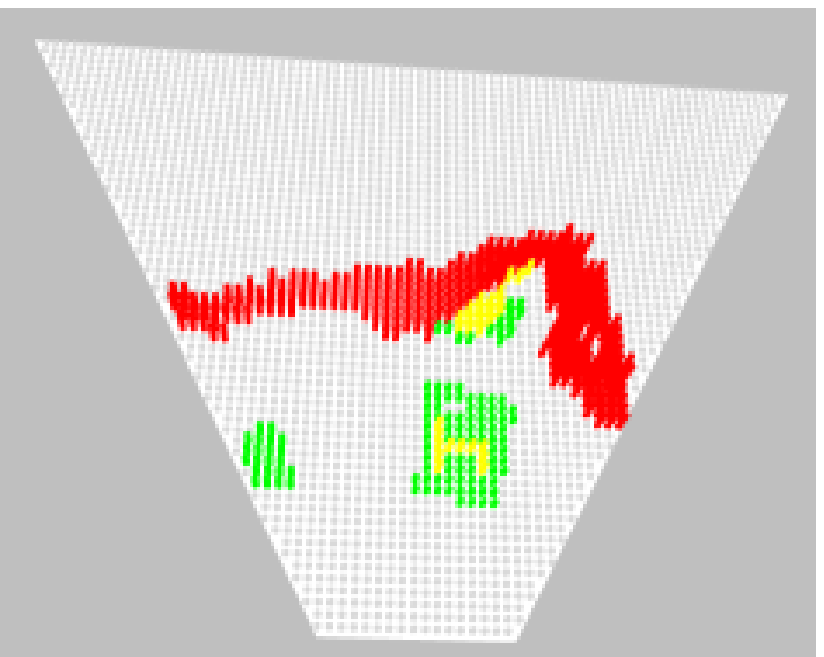}
		\includegraphics[width=2.7cm]{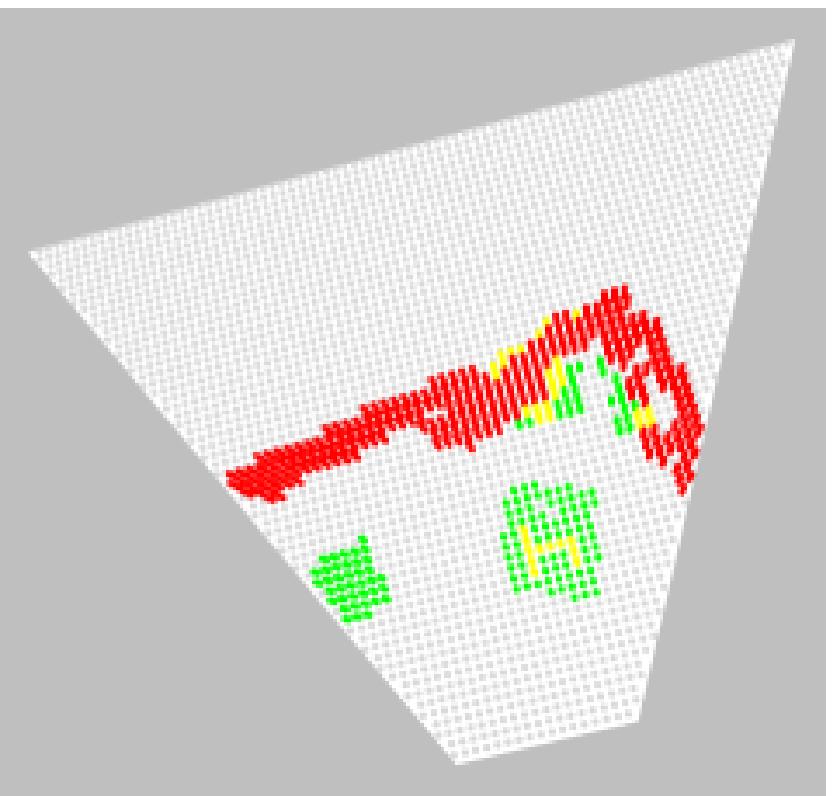}
		\includegraphics[width=2.7cm]{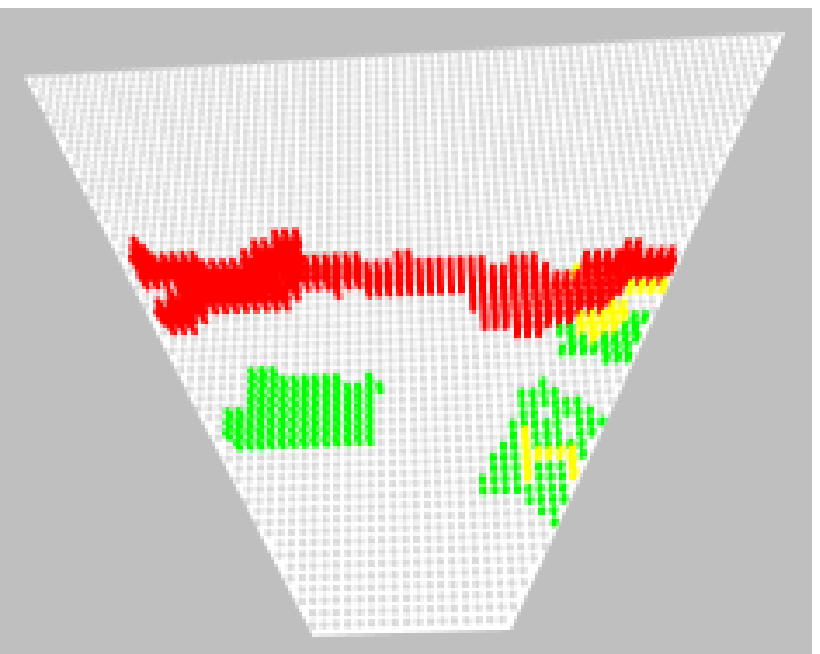}
		\includegraphics[width=2.7cm]{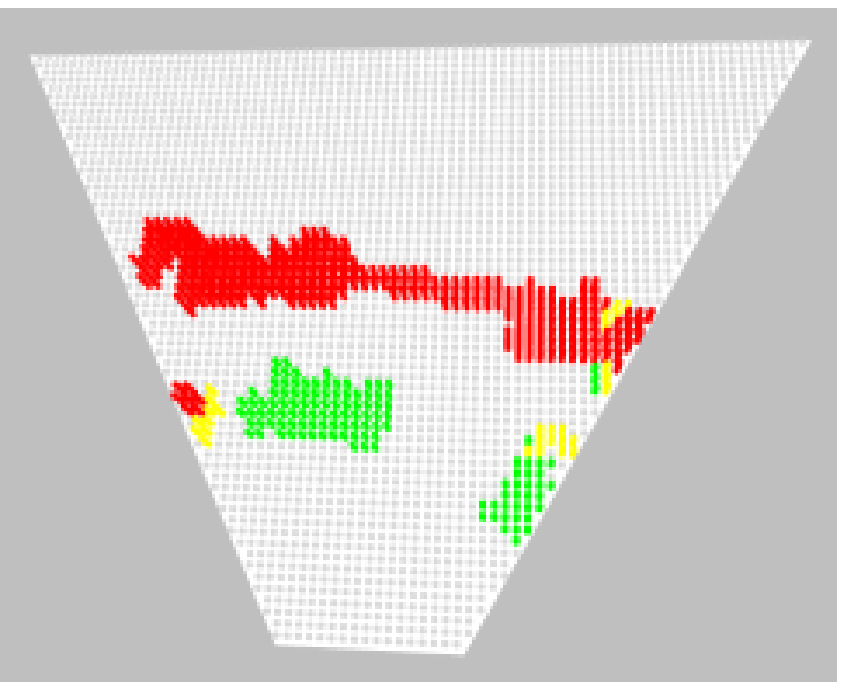}
		\caption{(From Top) The RGB image, the depth image, the output of the ground detection algorithm, the segmentation using DBSCAN and the semantic labeling mapped on the synthesis area.}
		\label{fig:test2}
	\end{figure}

For more experiments using the entire system on real-world data, we show and discuss the output of each module for 4 images taken from a video sequence from our GDIS dataset \cite{GDIS2019} (Figure \ref{fig:test2}). After the ground detection (Figure \ref{fig:test2} third row), we applied the DBSCAN clustering algorithm to break the occupied space into coarse segments (Figure \ref{fig:test2} fourth row). After that, we compute for each segment its geometric features and its class. The Pointnet network has classified well the chair except for the last frame (Figure \ref{fig:test2} last row, last column), it was classified as a table with a low probability (0.53), so the class was not mapped, only the convex hull was mapped. The model failed to predict the right class because the chair's point cloud was segmented into sub segments and this is due to the noisy nature of the Microsoft Kinect V1. If the class is not predicted, we set all the pins that represent the segment to the segment's level, rather than mapping the class to that level.
%\subsection{Discussion}
%The proposed system combines the deep learning features and the geometric features to provide for the user a scene description: for each segment, its label, location, shape, occupied space and its height are computed and mapped on the synthesis area.
%
%\begin{enumerate}
%\item The system detects 7 semantic classes that cover 16 different object classes. For each semantic class, we proposed a semantic label which is mapped on the center of the occupied space. Providing and only mapping coarse information is the first step in the visually impaired assistive systems for the scene description task. It provides a global scene description of the captured scene and how the objects are arranged in.
%\item The proposed system fails to predict the semantic class of some point clouds due to the nature of real-world data that is incomplete and cropped. Different cropped objects can be similar to some cropped objects belonging to different classes. As current solution, the model ignores the predicted class if this latter has been predicted with a small probability.
%\end{enumerate}
%%%%%%%%%%%%%%%%%%%%%%%%%%%%%%%%%%%%%%%%%%%%%%%%%%%%%%%%%%%%%%%%%%%%%%%%%%%%%%%%%%%%%%%%%%%%%%%%%%%%%%%
%						Future Scope
%%%%%%%%%%%%%%%%%%%%%%%%%%%%%%%%%%%%%%%%%%%%%%%%%%%%%%%%%%%%%%%%%%%%%%%%%%%%%%%%%%%%%%%%%%%%%%%%%%%%%%%
\section{Future Scope}\label{fcope}
In discussion section, we have mentioned two points that can be improved in the future respectively:
\begin{enumerate}
\item A second step can be added in the future for fine segmentation. In this step, objects that are arranged on a given coarse segment can be detected using coarse to fine segmentation. These latter, can be mapped on the synthesis area only by commands from the user to avoid ambiguity while exploring the scene. In other terms, mapping coarse and fine information at the same time is time consuming and ambiguous such as mapping the table and the objects that are on it. We can imagine this scenario: the user is searching for a cup. After providing the global description, the user can locate the table on the synthesis area. After that, by long clicking on the table's label, we can at this time map on the whole surface of the area only the table and what's on it. At this step, by touching the new area's configuration, the user can have an idea about what is on the table and can search for his desired item.

\item A solution that we will consider in the future, is that the proposed model will be extended to work for a succession of frames and thus, the cropped objects can be completed by the alignment process and thus the prediction can be corrected.
\end{enumerate}

On the other hand, we aim to cover other object classes. We also aim to provide a detailed semantic labeling, in a clear way, to include not only salient objects like tables, but also other small objects. This can be done by hierarchical classification by considering the classification of the salient objects. For example, to classify objects that are on the table the system will consider that the object is on the table.
%%%%%%%%%%%%%%%%%%%%%%%%%%%%%%%%%%%%%%%%%%%%%%%%%%%%%%%%%%%%%%%%%%%%%%%%%%%%%%%%%%%%%%%%%%%%%%%%%%%%%%%
%						CONCLUSION
%%%%%%%%%%%%%%%%%%%%%%%%%%%%%%%%%%%%%%%%%%%%%%%%%%%%%%%%%%%%%%%%%%%%%%%%%%%%%%%%%%%%%%%%%%%%%%%%%%%%%%%
\section{Conclusion} \label{conc}
In this paper, we proposed three main contributions:\\
-An object classification approach suitable for the visually impaired and blind people. It is based on deep learning networks implemented to classify the captured objects using a depth camera. Although we are using a state of the art neural network, we show that grouping object classes that have the same us- age improves the system accuracy by 5\%. This classification is combined with our previous approach \cite{zatout2019ego} in order to extend the number of possible classes and to provide additional information. Using only a depth camera as an input sensor reduces the computational complexity and the system's cost in terms of the required hardware.\\
-A semantic labeling that is similar to the Braille and Kanji systems, can be mapped into a synthesis area. The proposed classification approach and its adequate semantic labeling will allow the user to have a clear idea about his surroundings and thus, he can perform several tasks alone. \\
-An end-to-end assistive system for the visually impaired and blind people. This system describes the captured scene architecture and provides geometric features and the nature of the detected obstacles mapped as touch-based semantic labels. This latter will help the user to understand his surroundings
and thus, he can accomplish a set of tasks alone. 

The proposed system fails to predict the semantic class of some point clouds due to the nature of real-world data that is incomplete and cropped. Different cropped objects can be similar to some cropped objects belonging to different classes. As current solution, the model ignores the predicted class if this latter has been predicted with a small probability.
%%%%%%%%%%%%%%%%%%%%%%%%%%%%%%%%%%%%%%%%%%%%%%%%%%%%%%%%%%%%%%%%%%%%%%%%%%%%%%%%%%%%%%%%%%%%%%%%%%%%%%%
%						CONTENT ENDS
%%%%%%%%%%%%%%%%%%%%%%%%%%%%%%%%%%%%%%%%%%%%%%%%%%%%%%%%%%%%%%%%%%%%%%%%%%%%%%%%%%%%%%%%%%%%%%%%%%%%%%%
\section*{Acknowledgments}
{This work has been supported by the Algerian grant for PhD studies PRFU number C00L07UN160420190001. The authors acknowledge the financial support of the Directorate General for Scientific Research and Technological Development (DGRSDT) of Algeria for the Research Laboratory in Intelligent Informatics, Mathematics and Applications (RIIMA).
}
\section*{Compliance with Ethical Standards}
a)\textbf{Disclosure of potential conflicts of interest}\\
The authors Chayma Zatout and Slimane Larabi declare that they have no conflict of interest.\\
b)\textbf{Research grants from funding agencies}\\
This research was not funded.
\bibliographystyle{unsrt}  
%\bibliography{references}  %%% Remove comment to use the external .bib file (using bibtex).
%%% and comment out the ``thebibliography'' section.

%%% Comment out this section when you \bibliography{references} is enabled.

\end{document}